\documentclass[runningheads]{llncs}
\pdfoutput=1
\def\isfinal{}
\def\isextended{}
\usepackage{hyperref}
\usepackage[noend]{algpseudocode}
\usepackage{amsmath, amssymb, latexsym}
\usepackage[capitalise,noabbrev,nameinlink]{cleveref}
\usepackage{graphicx}
\usepackage{wrapfig}
\usepackage{pgfplots}
\usepackage{algorithm}
\usepackage{subcaption}

\usepackage{ifthen}
\ifthenelse{\isundefined{\isfinal}}{
    \newif\ifdebug\debugtrue
}{
    \newif\ifdebug\debugfalse
}
\ifthenelse{\isundefined{\isextended}}{
    \newif\ifextended\extendedfalse
}{
    \newif\ifextended\extendedtrue
}

\ifextended
\fi

\usepackage{tikz}
\usetikzlibrary{
arrows,
calc,
patterns,
}

\let\vec\mathbf

\newcommand{\enquote}[1]{``#1''}
\newcommand{\fakepar}[1]{\paragraph*{#1}}

\newcommand{\bbN}{\mathbb{N}}
\newcommand{\bbZ}{\mathbb{Z}}
\newcommand{\bbZi}{(\mathbb{Z}\uplus \{+\infty,-\infty\})}
\newcommand{\calC}{\mathcal{C}}

\newcommand{\size}[1]{\mathrm{size}(#1)}
\newcommand{\parts}[1]{\mathcal{P}(#1)}
\newcommand{\inter}[2]{[#1,#2]} 

\newcommand{\Cube}[2]{\mathrm{Cube}(#1,#2)}
\newcommand{\Corners}[1]{\mathrm{Corners}(#1)}
\newcommand{\MCorners}[1]{\overline{\mathrm{Corners}}(#1)}
\newcommand{\mCorners}[1]{\underline{\mathrm{Corners}}(#1)}
\newcommand{\ext}[2]{\mathop{\updownarrow}^{\nonscript#2}{#1}}
\newcommand{\ball}[1]{\mathcal{C}(#1)}

\newcommand{\oracle}[1]{\ensuremath{#1}}

\newcommand{\intb}{\mathcal{B}}

\tikzstyle{point}=[draw,circle,inner sep=1pt,fill,solid]
\tikzstyle{mcorner}=[point,inner sep=3pt, rectangle,anchor=south west,color=blue] 
\tikzstyle{Mcorner}=[mcorner,anchor=north east,color=blue!40!black] 
\tikzstyle{cex}=[point,color=red] 

\tikzstyle{box}=[fill=gray!30!white] 
\tikzstyle{learned}=[dashed, thick] 
\tikzstyle{del}=[pattern=north west lines] 
\tikzstyle{hyp}=[dotted] 

\newcommand{\OMIT}[1]{}
\newcommand{\Z}{\mathbb{Z}}
\newcommand{\concept}{X}
\newcommand{\conceptCl}{\mathcal{C}}
\newcommand{\baseSet}{\mathcal{D}}
\newcommand{\IntTh}{\langle \mathbb{Z};+\rangle}

\begin{document}

\title{Learning Union of Integer Hypercubes with Queries\ifextended
~--~Technical Report\fi\thanks{
This research was supported by the ERC Starting Grant 759969 (AV-SMP) and
Max-Planck Fellowship.
}
}
\subtitle{(with Applications to Monadic Decomposition)}

\author{
Oliver Markgraf\inst{1}\orcidID{0000-0003-4817-4563} \and
Daniel Stan\inst{1}\orcidID{0000-0002-4723-5742} \and
Anthony~W.~Lin\inst{1,2}\orcidID{0000-0003-4715-5096}
}

\institute{TU Kaiserslautern, Germany\\
\email{\{markgraf,stan,lin\}@cs.uni-kl.de}
\and
Max Planck Institute for Software Systems, Germany}

\maketitle

\begin{abstract}

We study the problem of learning a finite union of integer 
(axis-aligned) hypercubes over the $d$-dimensional integer lattice, i.e.,
whose edges are parallel to the coordinate axes. This is a natural
generalization of the classic problem in the computational learning theory of 
learning rectangles.
 We provide a learning algorithm with access to
a minimally adequate teacher (i.e. membership and equivalence oracles) that 
solves this problem in polynomial-time,
for any fixed dimension $d$. Over a non-fixed dimension, the problem subsumes
the problem of learning DNF boolean formulas, a central open problem in the
field. 
We have also provided extensions to handle infinite hypercubes in the union,
as well as showing how subset queries could improve the performance of the
learning algorithm in practice.
Our problem has a natural application to the problem of monadic 
decomposition of quantifier-free integer linear arithmetic formulas,
which has been actively studied in recent years. In particular, a finite union
of integer hypercubes correspond to a finite disjunction of monadic
predicates over integer linear arithmetic (without modulos constraints). 
Our experiments suggest that our learning algorithms substantially outperform 
the existing algorithms.

\end{abstract}

\section{Introduction} \label{sec:intro}

Suppose that we are interested in finding a formula $\varphi(\bar x)$ over some
theory $T$ (e.g. integer linear arithmetic) to ``capture'' a certain phenomenon,
which in verification could be, for instance, an invariant that a program 
satisfies some safety property. The
process of discovering $\varphi$ can be captured by the notion of a
\emph{learning algorithm} by allowing certain types of queries as an interface 
to some teacher \cite{Ang88}. Most standard learning frameworks can be captured
in this way. Here are some examples. Valiant's well-known notion of 
\emph{PAC-learning} can be captured by an oracle that returns a new random 
sample from an unknown distribution. Angluin's well-known notion of 
\emph{exact learning} \cite{Ang87,Ang88} can be captured by an interaction
with the so-called \emph{minimally adequate teachers}, which can answer
membership and equivalence queries. This has many applications in verification,
e.g., verification of parameterized systems \cite{CHLR17,LR16,Mar20} and
compositional verification \cite{Chen09}. Another learning framework
that has become very popular in verification is CEGIS (Counterexample Guided
Inductive Synthesis) \cite{Solar08,LMN16}, wherein a learning algorithm can ask 
equivalence
queries, but expect various types of ``constraint-like'' counterexamples 
(e.g. implication counterexamples) to be returned by the teacher. This is
of course in contrast to Angluin's exact learning setting, wherein the teacher 
may return only a positive/negative counterexample (a point in the symmetric 
difference of the target concept and the hypothesis).

In this paper, we study the problem of learning sets of points over the 
$d$-dimensional integer lattice that can be expressed as a \emph{finite union of
integer (axis-aligned, a.k.a. rectilinear) hypercubes}, i.e., whose edges are 
parallel to the
coordinate axes. Such a concept class of course forms a strict subclass of
sets of points that are definable by a formula $\varphi(x_1,\ldots,x_d)$
in the integer linear arithmetic (a.k.a. \emph{semi-linear sets}), which have
been addressed in several papers including \cite{Abe95,Takada92,KT10}, whose
PAC-learnability is as hard as PAC-learning boolean formulas in DNF
\cite{learning-book} ---
a long-standing open problem in learning theory ---
when binary representations are permitted (even over
dimension one \cite{Abe95}). That said, finite unions of integer hypercubes
are a concept class that naturally arises in computer science. 
Below we mention a few examples. 

The problem 
of learning rectangles (2-cube) and generalization to $d$-dimension are a 
classic example in computational learning theory, e.g., see 
\cite{learning-book,MT94}. Maass and Tur\'{a}n \cite{MT94} showed for example 
that the $d$-dimensional rectilinear cubes can be learned in polynomial-time 
with $O(\log n)$ queries, where the corners of the cubes are represented in 
binary. The authors posed as an open problem if one can learn a union of two 
(possibly overlapping) rectangles with only $O(\log n)$ equivalence queries. 
Chen \cite{C95} showed that this can be learned with 2 equivalence queries
and $O(d.\log n)$ membership queries. Later Chen and Ameur \cite{CA99} showed
that there is a polynomial-time algorithm using at most $O(\log^2 n)$ 
queries. The same paper left as an open problem if there is a
polynomial-time exact learning algorithm that learns finite unions of 
rectlinear cubes over a fixed dimension $d$. In this paper, \emph{we answer 
this in the positive}, and further show that this can be extended to allow 
\emph{infinite rectilinear hypercubes}, which in turn allow interesting 
applications in formal verification, as we discuss below.

Finite unions
of rectilinear cubes arise naturally in program analysis and verification.
Here we mention two examples. First, solving games over a large game graph 
has benefited from constraint-based
approaches, where winning regions can be succinctly represented and checked 
efficiently \cite{BCPR14}. For example, the discretization of the 
Cinderella-Stepmother problem \cite{BCPR14} admits winning regions that may
be represented by a union of a small number of cubes.
Secondly, verification algorithms benefit from optimization techniques like
	monadic decomposition~\cite{monadic-decomposition}, where the aim is the
    rewriting of a given quantifier-free SMT formula $\varphi(x_1,\ldots,x_n)$
    into an equivalent boolean combination of 
    monadic predicates $\psi(x_i)$ in some special form, i.e., typically in 
    DNF \cite{Libkin03,HLRW20,BHLLN19,CCG06},
    or by an if-then-else formula \cite{monadic-decomposition}, which could 
    sometimes be exponentially smaller than the DNF equivalent representation.
	Veanes \emph{et al.}\cite{monadic-decomposition} provided a generic semi-decision procedure for
	performing this monadic decomposition as an if-then-else formula, which 
    works regardless of the base theory.
	The restriction of the problem to the quantifier-free theory of integer 
	linear arithmetic (with and without extra modulo constraints) was studied in 
	\cite{HLRW20}, wherein the problem was shown to be
	coNP-complete and a monadic decomposition could be exponentially large in
	general. 
	For the subcase without modulo constraints, a monadic decomposition in DNF
    corresponds precisely to a finite union of (possibly infinite) rectilinear 
    hypercubes, which is the subject of this paper. 
    We describe below how oracles for memberships and equivalence (as well
    as more powerful queries like subsets) admit a fast implementation via an 
    SMT-solver, which enable our learning algorithms to be applied to compute 
    such a monadic decomposition.
    \OMIT{
    Since
    checking memberships (resp.~equivalence) corresponds to checking if 
    $\varphi(\bar c)$ is true which can be done in polynomial-time
    (resp.~$\forall \bar x( \varphi(\bar x) \leftrightarrow \psi(\bar x) )$,
    for a given quantifier-free formula $\psi$, which is coNP-complete, but
    can be checked by an optimized SMT-solver like Z3 \cite{z3}), a exact 
    learning algorithm for a finite union of (possibly infinite) rectilinear 
    hypercubes is applicable.

    \OMIT{
    , which
    effectively 
    our learning algorithm is able 
    to rewrite
	$\varphi$ to a monadic decomposition of $\varphi$ as
	a finite union of (possibly infinite) rectilinear hypercube.
}
    }

\OMIT{
, i.e., in each of those conjunction, the 
free variables are ``independent''.
Formally, each term consists in a Cartesian product of uni-variate constrains,
that we can relate here to our notion of cube.
}

\paragraph*{Contributions.}
We study the problem of learning finite unions of rectilinear hypercubes 
(over $\Z^d$) in 
Angluin's exact learning framework with membership and equivalence queries
\cite{Ang87,Ang88}. Our result is a polynomial-time exact learning algorithm 
for learning finite unions of rectilinear hypercubes over $\Z^d$ for fixed $d$.
This answers an open problem of \cite{CA99}. As observed in \cite{CA99}, 
over non-fixed $d$, this problem generalizes DNF since each term can be seen as
a hypercube over $\{0,1\}^d$. That is, without fixing $d$, the problem is
as hard as learning unrestricted DNF, which is well-known to be a major open 
problem in computational learning theory \cite{Angluin95}.

In view of applying our learning algorithm to the monadic decomposition problem
\cite{monadic-decomposition,HLRW20}
for quantifier-free integer linear arithmetic formulas, we consider two 
extensions. Firstly, we allow \emph{infinite hypercubes}. For example, over 
1-dimension, these would include infinite intervals like $[7,\infty)$, which
would correspond to the formula $x \geq 7$. Secondly, we observe that the 
\emph{subset query} (i.e. checking if the target concept includes a given finite
union $H$ of hypercubes) is not an expensive query for performing monadic
decomposition, i.e., it would correspond to a single satisfiability check of 
a quantifier-free integer linear arithmetic formula, which can be handled easily
by an SMT-solver. Subset queries belong to one of the standard types of
queries in Angluin's active learning framework, e.g., see \cite{Ang88}.
For this reason, we provide an optimization of our learning algorithm by means 
of subset 
queries.

We implemented these learning algorithms (vanilla flavour and 
various optimization including subset queries and 
``unary/binary acceleration''), using Z3 
\cite{z3} as the backend for answering equivalence and 
subset queries (each a satisfiability check of a quantifier-free 
formula). We have performed a micro-benchmarking to stress-test our algorithms 
against 
the generic monadic decomposition procedure of \cite{monadic-decomposition}, 
which also use Z3 as the backend, using various geometric objects over
$\mathbb{Z}^d$ as benchmarks. 
Our experiments suggest that our algorithms substantially outperform the
generic procedure.

\paragraph*{Organization.}
Preliminaries are in Section \ref{sec:prelim}. We present the
\emph{overshooting algorithm} that witnesses polynomial
learnability of finite unions of rectilinear cubes over a fixed dimension
$d$ with membership and equivalence in Section \ref{sec:exact}. In Section
\ref{sec:maximal}, we provide two extensions: (1) how subset queries could help
speed up the overshooting algorithm, (2) how the algorithm could be extended to
handle infinite cubes. Applications to monadic decomposition and experiments are
presented in Section \ref{sec:exp}. We conclude in Section \ref{sec:conclusion}.

We refer the reader
\ifextended
to the appendix
\else
to the technical report~\cite{techreport}
\fi
when proofs are omitted
and to the artifact~\cite{artifact} for implementation and benchmark details.

\paragraph{Acknowledgements.}
The authors would like to thank the reviewers and program chairs for their
thoughtful comments and suggestions to improve the presentation, as well as
Christoph Haase for fruitful discussions.

\section{Preliminaries}\label{sec:prelim}

We introduce below some common mathematical notations:
    $\bbN$ and $\bbZ$ are the sets of natural numbers and integers, 
    respectively.
    For $a,b\in\bbZ$, we write $\inter{a}{b} = \{i~|~a\leq i \leq b\}$;
    For any set $X$, we denote its power-set $\parts{X}$ and its cardinal
   $|X|\in \mathbb{N}\uplus\{\infty\}$;
    Given two sets $A,B$, the \emph{symmetric difference} is written
        $A\Delta B = A\backslash B \cup B\backslash A$;

    When analyzing complexity of the presented algorithms,
    we assume binary encoding for any number $n\in \bbZ$,
    which is part of the input of the considered algorithms, namely,
    $\size{n}=1+\lceil \log(|n|+1) \rceil$, where $\log$ is the base 2 logarithm. 

\fakepar{Hypercubes}
For a fixed \emph{dimension} $d\in\bbN$, we consider the \emph{discrete lattice}
$\bbZ^d$. A \emph{point} $\vec{v}\in\bbZ^d$ can be described by its coordinates
$\vec{v}[k]$ for $k\in \inter{1}{d}$. Let $\vec{v}[k/\alpha]$ denote the vector
$\vec{v}$ where the $i$-th coordinate has been replaced by $\alpha\in\bbZ$.
The notation $\vec{0}^d = (0,\ldots,0) \in \bbZ^d$ denotes the origin, or 
simply $\vec{0}$ when the dimension is clear from context.
We use standard notation for component-wise additions and scalar multiplication.
In particular, for $\alpha\in\bbZ$, $\vec{v}+\alpha\cdot \vec{v}'$ denotes the
vector $\vec{v}''\in\bbZ^d$ such that for all $i$, $\vec{v}''[i] = \vec{v}[i] + \alpha\cdot 
\vec{v}'[i]$. For $1\leq i \leq d$, we write $\vec{e}_i$ for the $i$-th 
\emph{elementary vector}, $\vec{e}_i = \vec{0}[i/1]$. We shall be mostly 
using the standard \emph{component-wise order} $\leq$ over vectors in $\Z^d$: 
$\vec{v}\leq \vec{v}'$ iff for all $i$, $\vec{v}[i] \leq \vec{v}'[i]$. We finally denote the
size of a vector as the sum of the sizes of its components:
$\size{\vec{v}} = \sum_{i=1}^d \size{\vec{v}[i]}$, for any $\vec{v}\in\bbZ^d$.

Our main study focuses on \emph{rectilinear hypercubes} (\emph{cubes} for 
short), i.e., any set of points
of the form $C = \{\vec{v}~|~\underline{\vec{v}} \leq \vec{v} \leq \overline{\vec{v}}\}$
for some $\underline{\vec{v}}\leq \overline{\vec{v}}\in \bbZ^d$.
The size of $C$ is uniquely defined as
$\size{C}=\size{\underline{\vec{v}}} + \size{\overline{\vec{v}}}$.
On the contrary, an arbitrary finite set $\concept$ has no
unique representation as a finite union of cubes,
therefore we define its size as the size of its best representation:
\[
  \size{\concept} = \min\left\{
    \sum_{i=1}^n \size{\underline{\vec{v}}_i} + \size{\overline{\vec{v}}_i} ~\middle|~
    \exists n, \underline{\vec{v}}_1 \hdots \overline{\vec{v}}_n:
    \concept = \bigcup_{i=1}^n \Cube{\underline{\vec{v}}_i}{\overline{\vec{v}}_i}
  \right\}
\]
We adopt here a worst-case analysis approach, where our later reasoning and
complexity analysis are valid for any representation,
they are in particular valid for its best representation.

\fakepar{Learning model}
We first recall some standard definition from computational learning theory;
for more, see \cite{learning-book}. Fix a countable \emph{base set} $\baseSet = 
\bigcup_{i=1}^n 
\baseSet_i$, where the sets $\baseSet_i$'s are pairwise disjoint. The problem
of learning boolean formulas in DNF uses $\baseSet_i = \{0,1\}^i$, i.e.,
the set of all binary sequences of length $i$, which can be thought of as
a set of all assignments to a boolean function over $x_1,\ldots,x_i$.
The learning problem in this paper uses $\baseSet_i = \Z^i$.
A \emph{concept} $\concept$ is simply a subset of $\baseSet_i$, for some $i
\in\Z_{> 0}$. For example, when $\baseSet_i = \{0,1\}^i$, a concept is simply
a boolean function over $x_1,\ldots,x_i$.
When we speak of a learning problem, we always have a fixed set of
representations in mind. For example, when we speak of learning boolean formulas
in DNF (Disjunctive Normal Form), the representation $\varphi_\concept$ of a 
boolean function $\concept$ has to be a formula over $x_1,\ldots,x_i$ in DNF.
For example, 
$\concept$ could be a boolean function, whereas $\varphi_{\concept}$ a DNF
formula
representing $\concept$. Note that a concept could admit many possible 
representations.
A \emph{concept class} $\conceptCl = \bigcup_{i=1}^\infty \conceptCl_i$ is a 
set of concepts, where $\conceptCl_i \subseteq \parts{\baseSet_i}$. 
For example,
$\conceptCl_i$ could be the set of boolean functions over variables
$x_1,\ldots,x_i$. When the set of representations for $\conceptCl$ is fixed
(e.g. DNF for representing boolean functions), we could define 
$\size{\concept}$ of the concept $\concept$ to be the size of the smallest 
representation of $\concept$. In this paper, we are dealing with the concept
class $\mathcal{C}_d \subseteq \parts{\Z^d}$ of sets of integer points that
can be represented as a finite union of rectilinear hypercubes over $\Z^d$.
Earlier  in this section we have defined this concept, as well as the size
of the representation. To avoid notational clutter, we will often denote the
concept class $\conceptCl_d$ by $\conceptCl$ because our algorithm typically 
assumes that $d$ is fixed.

In Angluin's active learning framework \cite{Ang87,Ang88}, the learner has
access to oracles (a.k.a. teachers) that could provide hints about the target 
concept $\concept$
to the learner. A \emph{minimally adequate teacher} must be able to answer membership
and equivalence queries.
\OMIT{
here some important notions of the learning framework, in a proper
setting\cite{Ang87,Ang88}. We focus on learning set of points of a given
representation. This class of sets of interest will later be called
\emph{concept class}, usually written $\calC$,
and denote any countable set $\calC\subseteq \parts{\bbZ^d}$.
}
\begin{definition}[M+EQ Oracles]
    \label{def:meqoracles}
    Consider some target concept $\concept\in \conceptCl_d$ for some concept 
    class $\calC = \bigcup_{d=1}^\infty \conceptCl_d$
    and let $\bot,\top\notin \baseSet$ be two fresh symbols.
    \begin{itemize}
       \item A \emph{membership oracle} (\oracle{M}) for $\concept$ is a 
        function
        $\Phi_\concept: \baseSet_d \rightarrow \{ \top, \bot \}$, which outputs 
        $\top$ iff $\vec{v} \in \concept$.
        \item An \emph{equivalence oracle} (\oracle{EQ}) for $\concept$ is a 
            function
            $\Psi_\concept: \calC_d \rightarrow \baseSet_d \uplus \{\top\}$
            such that for all \emph{hypothesis} $H\in\calC$, 
            $\Psi_\concept(H)\in (H\Delta \concept)\uplus\{\top\}$
            and
            $\Psi_\concept(H) = \top$ implies $H=\concept$.
    \end{itemize}
\end{definition}
Intuitively, an equivalence oracle tells, for any hypothesis $H\in\calC$,
whether $H=\concept$. If yes, $\top$ is returned; if not, it provides a 
\emph{counterexample}, namely a point in the symmetric difference.
Angluin has considered other types of queries as well in her framework including
subset/superset queries and difference queries (e.g. see her excellent survey
\cite{Ang88}). We will use the subset queries in Section \ref{sec:maximal}.

A learning algorithm $\mathcal{A}$ is said to \emph{learn} the concept class 
$\conceptCl = \bigcup_{d=1}^\infty \conceptCl_d$ if, given $d$ as input and
any unknown target concept $\concept$, it terminates and outputs a 
representation of $\concept$ after a finite amount of interaction with the
oracles. Assuming that the oracle always returns the shortest counterexamples, 
its running time is defined to be number of steps (measured in $d$ and 
$\size{\concept}$) that $\mathcal{A}$ takes to output a representation of 
$\concept$. The complexity $comp(d,\size{\concept})$ of $\mathcal{A}$
measures the number of steps taken in the worst case for all $d$ and
$\size{\concept}$. It runs in polynomial time if $comp$ is a polynomial
function. It remains a long-standing open problem in computational learning 
theory if there is a learning algorithm for boolean formulas represented in DNF,
which is true for almost all major models including exact learning and PAC
(see \cite{Angluin95}). Over geometric concepts including hypercubes and
semilinear sets, the dimension $d$ is sometimes considered a fixed parameter, 
e.g., see \cite{MT94,Abe95,CA99,KT10}.

\OMIT{
Provided a set of oracles for a target set $\concept\in\calC$, we are interested
in building an algorithm making a finite number of \emph{queries}, that is
to say evaluating the oracle functions on some input (point or hypothesis),
and terminating with a representation of~$\concept$.
}

\section{Minimally adequate teacher}
\label{sec:exact}

We restrict first our attention to the minimally adequate teacher setting where
only a membership and equivalence oracle are provided, and provide constructions
for intermediate procedures that can be interpreted as oracles.

\subsection{Corner oracle}
\label{ssec:corneroracle}
At the heart of our learning algorithm is the concept of corners:
\begin{definition}
Given a set of points $X\subseteq \bbZ^d$,
a \emph{maximal corner} (resp \emph{minimal corner}) of $X$ is a point $\vec{v}\in X$
maximal (resp minimal) with respect to component-wise ordering $\leq$.
We write $\MCorners{X}$ and $\mCorners{X}$ for the sets of maximal and minimal
corners, respectively, and write
$\Corners{X}=\MCorners{X}\cup \mCorners{X}$.
\end{definition}

Given a membership oracle for some $X\in \calC$ containing $\vec{0}$,
Algorithm \ref{alg:findMaxCorner} returns \emph{some} maximal corner of a given 
finite subset.
\begin{algorithm}
\begin{algorithmic}
  \Ensure Returned value is a maximal corner of~$X$
  \Require $\vec{0}\in X;~ \Phi_X$~a membership oracle for $X$
  \Function{findMaxCorner}{$\Phi_X$}
    \State $i \gets 0;\quad \vec{v} = \vec{0}$
    \While{$i<d$}
      \State $i \gets i+1; \quad k \gets 1;\quad l \gets 1$;
      \If{$\Phi_X(\vec{v}+\vec{e}_i)$}
        \While{$\Phi_X(\vec{v}+k\cdot \vec{e}_i)$}
           \State $l\gets k; \quad k\gets 2k$
        \EndWhile
        \While{$k-l > 1$}
          \If{$\Phi_X(\vec{v}+\lfloor(k+l)/2\rfloor\cdot e_i)$}
              \State $l\gets \lfloor(k+l)/2\rfloor$
          \Else
              \State $k\gets \lfloor(k+l)/2\rfloor$
          \EndIf
        \EndWhile
        \State $\vec{v}\gets \vec{v}+l\cdot \vec{e}_i;\quad i\gets 0$
       \EndIf
    \EndWhile
    \Return $\vec{v}$
  \EndFunction
\end{algorithmic}
\caption{Binary search for a maximal corner, assuming $\vec{0}\in X$}
\label{alg:findMaxCorner}
\end{algorithm}
Intuitively, for each coordinate $i$, a binary search is made until
a border of $X$ is eventually found. More precisely, we provide the following
complexity analysis.
\begin{proposition}
    \label{prop:membership}
    Let $\Phi_X$ be a membership oracle for
    $X=\cup_{i=1}^n \Cube{\underline{\vec{v}}_i}{\overline{\vec{v}}_i}$
    and assume $\vec{0} \in X$.
    Then $\Call{findMaxCorner}{\Phi_X}$
    terminates after
    $O\left(\sum_{j=1}^n \size{\overline{\vec{v}}_j}\right)$
    queries and returns some $\overline{\vec{v}} \in \MCorners{X}$.
\end{proposition}

This algorithm provides a partial implementation of the following
oracle:
\begin{definition}
  \label{def:corneroracle}
  Given $X\in\calC$, a \emph{corner oracle} for $X$ is any function
  $\Theta_X: X\rightarrow \mCorners{X}\times \MCorners{X}$.
\end{definition}

A complete implementation of this oracle is provided by noticing that membership
oracles can easily be composed:
\begin{remark}
   \label{rem:oraclecomp}
   Assume $\Phi_A$ and $\Phi_B$ are two given membership oracles,
   respectively for two arbitrary sets $A$ and $B$, and $f: \bbZ^d \rightarrow \bbZ^d$.
   One can build membership
   oracles for $A\cup B$, $A\cap B$, $A\Delta B$, $A\backslash B$
   and $f(A)$.
   In particular:
   \begin{itemize}
      \item By instantiating $f: \vec{v} \mapsto -\vec{v}$, the previous
        procedure applied on $\Phi_{f(A)}$
        returns some $\vec{v}\in \MCorners{-A}$, so $-\vec{v}\in\mCorners{A}$.
      \item For any $\vec{v}_0\in A$ and $f: \vec{v}\mapsto \vec{v}-\vec{v}_0$, 
   $\Phi_{f(A)}$ is a membership oracle for $A-\vec{v}_0 = \{\vec{v}~|~\vec{v}+\vec{v}_0\in A\}$
      containing $\vec{0}$, so
      \Call{findMaxCorner}{$\Phi_{f(A)}$}
      returns some $\vec{v}\in\MCorners{A-\vec{v}_0}$ so $\vec{v}+\vec{v}_0\in\MCorners{A}$.
   \end{itemize}
   In both cases, notice that $\size{f(A)} \leq \size{A}+\size{\vec{v}_0} \leq 2\size{A}$.
\end{remark}

In the sequel
we write $\Phi_C$ for the membership oracle of any set $C$ obtained by
composing sets whose oracles are provided.
We also assume having constructed the two procedures
$\Call{findMaxCorner}{\vec{v},\Phi_X}$ and 
$\Call{findMinCorner}{\vec{v},\Phi_X}$.

\subsection{Overshooting algorithm}
\label{ssec:overshooting}

\begin{algorithm}
\begin{algorithmic}
\Require $\Phi_X$ membership oracle for $X$, $\Psi_X$ equivalence oracle for $X$
\begin{minipage}{0.45\textwidth}
  \Function{LearnCubes}{$\Phi_X$,$\Psi_X$}
  \State $H \gets \emptyset$
  \Repeat
    \State $\vec{v}\gets \Psi_X(H)$
    \State $H \gets \Call{Refine}{H,\vec{v},\Phi_X}$
  \Until{$\vec{v} = \top$}
  \EndFunction
  \Function{RefineSym}{$H,\vec{v},\Phi_X$} 
    \State $\underline{\vec{v}} \gets \Call{findMinCorner}{\vec{v},\Phi_{X \Delta H}}$
    \State $\overline{\vec{v}} \gets  \Call{findMaxCorner}{\vec{v},\Phi_{X \Delta H}}$
    \State \textbf{return} $H \Delta \Cube{\underline{\vec{v}}}{\overline{\vec{v}}}$
  \EndFunction
\end{minipage}
\begin{minipage}{0.50\textwidth}
  \Function{RefineAddRemove}{$H,\vec{v},\Phi_X$} 
  \If{$\Phi_X(\vec{v})$}
    \State $\underline{\vec{v}} \gets \Call{findMinCorner}{\vec{v},\Phi_{X \backslash H}}$
    \State $\overline{\vec{v}}  \gets  \Call{findMaxCorner}{\vec{v},\Phi_{X \backslash H}}$
    \State \textbf{return} $H \cup \Cube{\underline{\vec{v}}}{\overline{\vec{v}}}$
  \Else
    \State $\underline{\vec{v}} \gets \Call{findMinCorner}{\vec{v},\Phi_{H\backslash X}}$
    \State $\overline{\vec{v}} \gets  \Call{findMaxCorner}{\vec{v},\Phi_{H\backslash X}}$
    \State \textbf{return} $H \backslash \Cube{\underline{\vec{v}}}{\overline{\vec{v}}}$
  \EndIf
  \EndFunction
\end{minipage}
\end{algorithmic}
\caption{Overshooting algorithms}
\label{alg:naive}
\end{algorithm}
The core loop of the learning algorithm is presented in
the \Call{LearnCubes}{} function of~\cref{alg:naive}. The hypothesis
is initially empty, and is later refined, as long as a counterexample is returned.
How to refine the hypothesis given a counterexample? Two implementations
of \Call{Refine}{} are provided namely \Call{RefineSym}{} and
\Call{RefineAddRemove}{}, giving rise to two variants of the algorithm.
In both cases, the refinement takes a counterexample as an input and uses
the corner oracle to build a cube~$C$. In the former variant, a symmetric
difference between the current hypothesis and $C$ is made, while in the latter,
$C$ is either added or removed from the hypothesis.

\begin{figure}
\centering
    \def\ax{1}\def\ay{1}
    \def\bx{3.5}\def\by{1.9}
    \def\cx{3}\def\cy{1.5}
    \def\dx{4}\def\dy{2.5}

    \def\shift{-2cm}
\begin{subfigure}[b]{0.32\textwidth}
  \begin{tikzpicture}
      \draw[box] (\ax,\ay) -- (\bx,\ay) -- (\bx,\cy) -- (\dx,\cy) -- (\dx,\dy)
      -- (\cx,\dy) -- (\cx,\by) -- (\ax,\by) -- (\ax,\ay);
      
      \node[cex] at ($(\ax,\cy)!0.5!(\dx,\by)$) {};
      \draw[learned] (\ax,\ay) node[mcorner] {}
          -- (\ax,\dy) -- (\dx,\dy) node[Mcorner] {} -- (\dx,\ay) -- (\ax,\ay);
      \begin{scope}[yshift=\shift]
      \draw[hyp] (\ax,\ay) 
          -- (\ax,\dy) -- (\dx,\dy) -- (\dx,\ay) -- (\ax,\ay);
      \end{scope}
  \end{tikzpicture}
  \caption{Step 1: add}
  \label{sfig:run1.1}
\end{subfigure}
\begin{subfigure}[b]{0.32\textwidth}
 \begin{tikzpicture}
     \draw[box] (\ax,\by) rectangle (\cx,\dy);
     \draw[box] (\bx,\ay) rectangle (\dx,\cy);
     \node[cex] at ($(\ax,\by)!0.5!(\cx,\dy)$) {};
     \draw[learned,del] (\ax,\by) node[mcorner] {} -- (\ax,\dy)
                 -- (\cx,\dy) node[Mcorner] {} -- (\cx,\by) -- (\ax,\by);
      \begin{scope}[yshift=\shift]
      \draw[hyp] (\ax,\ay) -- (\ax,\by) -- (\cx,\by) -- (\cx,\dy) -- (\dx,\dy)
        -- (\dx,\ay) -- (\ax,\ay);
      \end{scope}
 \end{tikzpicture}
  \caption{Step 2: remove}
  \label{sfig:run1.2}
\end{subfigure}
\begin{subfigure}[b]{0.32\textwidth}
  \begin{tikzpicture}
    \begin{scope}
      \node at (\ax,\ay) {};
     \draw[box] (\bx,\ay) rectangle (\dx,\cy);
      \node[cex] at ($(\bx,\ay)!0.5!(\dx,\cy)$) {};
      \draw[learned,del] (\bx,\ay) node[mcorner] {} -- (\bx,\cy)
                  -- (\dx,\cy) node[Mcorner] {} -- (\dx,\ay) -- (\bx,\ay);
      \begin{scope}[yshift=\shift]
      \draw[hyp] (\ax,\ay) -- (\bx,\ay) -- (\bx,\cy) -- (\dx,\cy) -- (\dx,\dy)
      -- (\cx,\dy) -- (\cx,\by) -- (\ax,\by) -- (\ax,\ay);
      \end{scope}
    \end{scope}

  \end{tikzpicture}
  \caption{Step 3: remove}
  \label{sfig:run1.3}
\end{subfigure}
\begin{tikzpicture}[every node/.style={node distance=1.5cm}]
    \tikzset{ann/.style={node distance=0.1cm, right of=a, anchor=west}}
    \node[cex] (fa) {};
    \node[ann, right of=fa, anchor=west] (a) {counterexample $v$};
    \node[mcorner, right of=a,node distance=2.2cm] (a) {};
    \node[ann] (a) {minimal corner $\underline{v}$};
    \node[Mcorner, right of=a, node distance=2.2cm] (a) {};
    \node[ann] (a) {maximal corner $\overline{v}$};
    
\end{tikzpicture}
\begin{tikzpicture}[every node/.style={node distance=1.5cm}]
    \tikzset{ann/.style={node distance=0.1cm, right of=a, anchor=west}}
    \node[box,inner sep=0.1cm,draw] (a) {};
    \node[ann] (a) {search space};

    \node[hyp,inner sep=0.1cm, right of=a,draw] (a) {};
    \node[ann] (a) {hypothesis};

    \node[learned,inner sep=0.1cm, right of=a,draw] (a) {};
    \node[ann] (a) {learned cube to add};

    \node[learned, del,inner sep=0.1cm, right of=a, anchor=west, draw] (a) {};
    \node[ann] (a) {learned cube to remove};
\end{tikzpicture}
\caption{Possible run of the overshooting algorithm on two cubes in $2$ dimensions}
\label{fig:run1}
\end{figure}
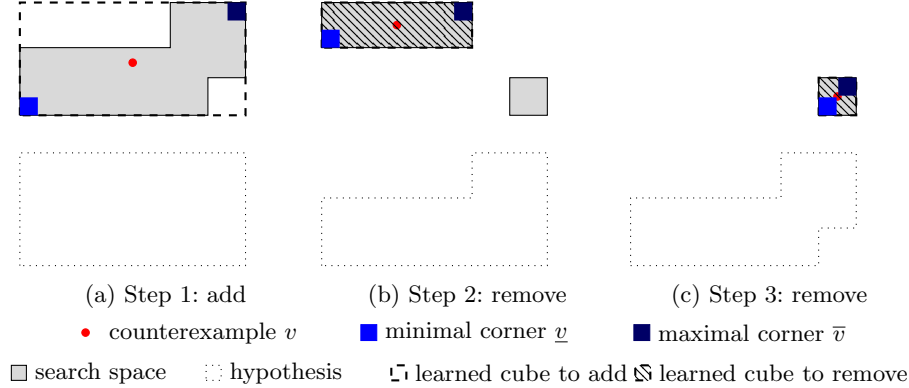

An example run of the \Call{RefineAddRemove}{} variant is depicted in~\cref{fig:run1}.
While the above diagrams represent the search space used by the corner oracles,
the below diagrams depict the resulting hypothesis after refinement.
Initially, the hypothesis is empty (not represented) so the search space
coincides with the target set~$X$, which can be represented as a union of two
overlapping cubes.
A counterexample $\vec{v}\in X\backslash H$ is
therefore returned by the equivalence oracle. As $\vec{v}\in X$,
the refinement procedure adds some cube by searching the state space
$X\backslash H=X$ around~$\vec{v}$. A too large cube is then added to the hypothesis,
and a negative counterexample $\vec{v}\in H\backslash X$ is then returned. The search
space is now $H\backslash X$ and the algorithm aims at removing some smaller
cube from the hypothesis. After two removals, the final hypothesis coincides with
the target.

\fakepar{Hypothesis representation}
Both variants are operating on the hypothesis by applying boolean operations.
One can naturally wonder if hypothesis represented by union, symmetric
differences and differences of cubes can be handled by oracles operating on
the concept class of finite cubes. As a matter of fact, we will observe that
$H\Delta X$, $H\backslash X$ and $X \backslash H$ can all be represented 
in $\calC$:

\begin{lemma}[Cube intersection and subtraction]
    \label{lem:cubsub}
    Let $C_1=\Cube{\underline{\vec{v}}_1}{\overline{\vec{v}}_1}$
    and $C_2=\Cube{\underline{\vec{v}}_2}{\overline{\vec{v}}_2}$ two cubes.

    Then $C_1 \cap C_2$ is a cube and $C_2 \backslash C_1$
        can be written as the disjoint union of $2d$ cubes.
        Moreover, these computations are effective in $2d$ operations.
\end{lemma}
Intuitively, one can think of a cube subtracted by a smaller cube results in
a family of cubes, one for each face of the larger cube. There are $2d$ faces
for a cube in dimension $d$.

\subsection{Repetition-free complexity}
  \label{ssec:compovershoot}
  In order to analyze the complexity of both variants of the algorithm, we fix a
  finite target set $X\in\conceptCl_d$ and one of its representation as a union
  of cubes:
        \[
            X = \bigcup_{i=1}^{n} \Cube{\underline{\vec{v}}_i}{\overline{\vec{v}}_i}
        \]
  
  We prove by induction on the iteration step that $H$ can be expressed
  as a union of cubes, whose corners are aligned on a particular set of
  points:
  \begin{definition}[Abstract grid]
    For $1\leq k \leq d$, we define the sets:
  \[
  \begin{aligned}
    \underline{B}_{k} =  \{
        \overline{\vec{v}}_i[k]+1~|~1\leq i\leq C
    \}
    \cup \{
        \underline{\vec{v}}_i[k]~|~1\leq i\leq C
    \}\\
    \overline{B}_{k} =  \{
        \overline{\vec{v}}_i[k]~|~1\leq i\leq C
    \}
    \cup \{
        \underline{\vec{v}}_i[k]-1~|~1\leq i\leq C
    \}
  \end{aligned}
  \]
     For any $A\subseteq \bbZ^d$, we write $A\in \intb$ whenever is a
     finite union of cubes of the form
        $\Cube{\vec{v}}{\vec{v}'}$ such that for all $k$,
        $\vec{v}[k] \in \underline{B}_k$ and
        $\vec{v}'[k] \in \overline{B}_k$.
  \end{definition}
  Intuitively, $\underline{B}_k$ (resp $\overline{B}_k$) describes all the
  possible $k$-coordinate for minimal corners (resp maximal). A coordinate
  for a max corner, i.e. a constraint of the form $x_k \leq \alpha$,
  can become a coordinate for a minimal corner, i.e. a constraint of the
  form $x_k \geq \alpha+1$, when taking
  the complement during a difference operation, and vice versa.

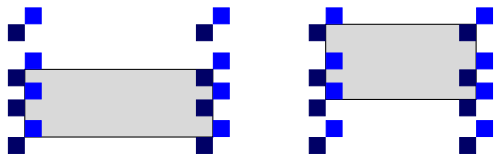
\begin{figure}
    \centering
    \def\ax{1}\def\ay{1}
    \def\bx{3.5}\def\by{1.9}
    \def\cx{5}\def\cy{1.5}
    \def\dx{7}\def\dy{2.5}
    \begin{tikzpicture}
       \draw[box] (\ax,\ay) rectangle (\bx,\by);
       \draw[box] (\cx,\cy) rectangle (\dx,\dy);
       \node[mcorner] at (\ax,\ay) {};
       \node[mcorner] at (\bx,\by) {};
       \node[mcorner] at (\cx,\cy) {};
       \node[mcorner] at (\dx,\dy) {};
       \node[mcorner] at (\ax,\by) {};
       \node[mcorner] at (\ax,\cy) {};
       \node[mcorner] at (\ax,\dy) {};
       \node[mcorner] at (\bx,\ay) {};
       \node[mcorner] at (\bx,\cy) {};
       \node[mcorner] at (\bx,\dy) {};
       \node[mcorner] at (\cx,\ay) {};
       \node[mcorner] at (\cx,\by) {};
       \node[mcorner] at (\cx,\dy) {};
       \node[mcorner] at (\dx,\ay) {};
       \node[mcorner] at (\dx,\by) {};
       \node[mcorner] at (\dx,\cy) {};

       \node[Mcorner] at (\ax,\ay) {};
       \node[Mcorner] at (\bx,\by) {};
       \node[Mcorner] at (\cx,\cy) {};
       \node[Mcorner] at (\dx,\dy) {};
       \node[Mcorner] at (\ax,\by) {};
       \node[Mcorner] at (\ax,\cy) {};
       \node[Mcorner] at (\ax,\dy) {};
       \node[Mcorner] at (\bx,\ay) {};
       \node[Mcorner] at (\bx,\cy) {};
       \node[Mcorner] at (\bx,\dy) {};
       \node[Mcorner] at (\cx,\ay) {};
       \node[Mcorner] at (\cx,\by) {};
       \node[Mcorner] at (\cx,\dy) {};
       \node[Mcorner] at (\dx,\ay) {};
       \node[Mcorner] at (\dx,\by) {};
       \node[Mcorner] at (\dx,\cy) {};
    \end{tikzpicture}
    \caption{Possible minimal and maximal corners for cubes appearing
        in the hypothesis, for a given target space}
    \label{fig:abstractgrid}
  \end{figure}

  We observe that $\intb$ is stable by union, intersection and difference.
  In particular, the overshooting algorithms maintain $H\in\intb$, namely
  the hypothesis always has minimal (resp maximal) corners
  that align with $\underline{B}_k$ (resp $\overline{B}_k$) on the $k$-th
  coordinate. \Cref{fig:abstractgrid} provides an example of
  such points for a target made of the union of two cubes.

  Since the sets $\underline{B}_k$ and $\overline{B}_k$ are of size at
  most~$2n$ for every~$k$, there are at most $(2n)^{2d}$ possible cubes, polynomial for a
  fixed~$d$. Assuming $H\in \intb$, we can ensure that \cref{lem:cubsub}
  maintains a polynomial representation of the hypothesis throughout the algorithm
  until termination.

\begin{figure}
\centering
  \begin{tikzpicture}
    \def\ay{0}
    \def\by{-1}
    \def\cy{-2}
    \def\h{0.5}
    \def\w{1}
    \def\spacing{2cm}

    \begin{scope}[xshift=-1cm]
      \draw[box] (0,\ay) rectangle +(\w,\h);
      \node[anchor=south west] at (0,\ay) {A};
      \draw[box] (0,\by) rectangle +(\w,\h);
      \node[anchor=south west] at (0,\by) {B};
      \draw[box] (0,\cy) rectangle +(\w,\h);
      \node[anchor=south west] at (0,\cy) {C};
    \end{scope}

    \begin{scope}[xshift=0.5cm]
        \node at ($(\w/2,\h/2)$) {$\emptyset$};
        \path (0,\by) node[mcorner] {}
            +(\w,\h) node[Mcorner] {};
        \node[cex] at ($(\w,\by)!0.5!(0,-\h)$) {};
    \end{scope}

    \begin{scope}[xshift=\spacing]
        \draw[hyp] (0,\by) rectangle +(\w,\h);
        \node[cex] at ($(0,0)!0.5!(\w,\ay)$) {};
        \node[Mcorner] at (\w,\h) {};
        \node[mcorner] at (0,\cy) {};
    \end{scope}

    \begin{scope}[xshift=2*\spacing]
        \draw[hyp] ($(0,\ay+\h)$) rectangle ($(\w,\cy)$);
        \path (0,\cy+\h) node[mcorner] {}
            (\w,\ay) node[Mcorner] {};
        \node[cex] at ($(0,\by)!0.7!(\w,\ay)$) {};
    \end{scope}

    \begin{scope}[xshift=3*\spacing]
        \draw[hyp] (0,\ay) rectangle +(\w,\h);
        \draw[hyp] (0,\cy) rectangle +(\w,\h);
        \path (0,\by) node[mcorner] {}
            +(\w,\h) node[Mcorner] {};
        \node[cex] at ($(\w,\by)!0.5!(0,-\h)$) {};
    \end{scope}

    \begin{scope}[xshift=4*\spacing]
        \draw[hyp] (0,\ay) rectangle +(\w,\h);
        \draw[hyp] (0,\by) rectangle +(\w,\h);
        \draw[hyp] (0,\cy) rectangle +(\w,\h);
    \end{scope}
    
  \end{tikzpicture}
\caption{Possible run on three cubes where cube B is added twice to the hypothesis.}
\label{fig:redundant}
\end{figure}
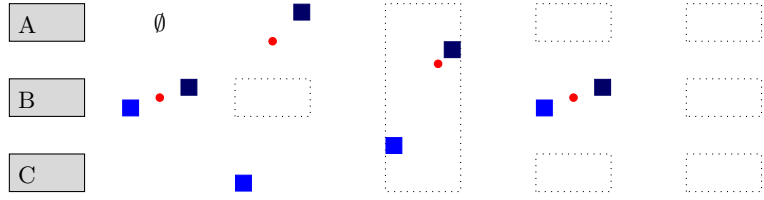

  Although $\intb$ is of polynomial size, proving $H\in\intb$ is not sufficient
  to prove termination of the algorithm in polynomial time, especially if
  some cubes in $\intb$ are added and removed several times. Consider for example
  \cref{fig:redundant} which depicts a possible run of the algorithm on three
  aligned cubes by its successive hypotheses: cube $B$ is added during the
  first step, but is later covered when the algorithm tries to learn $A$
  but overshoots. Another overshooting happens when trying to remove
  the space between $A$ and $B$, which ends up removing all space between $A$
  and $C$. The cube $C$ has then to be learned a second time, terminating
  the~algorithm.

To circumvent this issue, we propose an optimization that prevents
visiting twice the same minimal corner~$\underline{\vec{v}}$.
We base our reasoning on the following observations:
\begin{itemize}
  \item If $\vec{v}\in X$, then $\underline{\vec{v}}\in X$, so $\underline{\vec{v}}$
    should not be later removed.
  \item If $\vec{v}\notin X$, then $\underline{\vec{v}} \notin X$, so $\underline{\vec{v}}$
    should not be later~added back to~$H$.
\end{itemize}

\Cref{alg:opt} introduces an optimized refinement procedure to keep track of
the already added maximal corners.
Although an analogous optimization can be done on the symmetric difference variant,
we only discuss here~\Call{RefineAddRemove2}{}.

Once a minimal corner $\underline{\vec{v}}$ for a candidate cube has been
found, we continue the search of a maximal corner $\overline{\vec{v}}$
by avoiding points that will result in the removal
(resp addition) of already added (resp removed) minimal corners.
\begin{algorithm}
\begin{algorithmic}
  \State Let $V\gets \emptyset$
  \Function{RefineAddRemove2}{$H,\vec{v}_e,\Phi_X$} 
  \If{$\Phi_X(\vec{v}_e)$}
    \State Let $\underline{\vec{v}} = \Call{findMinCorner}{\vec{v}_e,\Phi_{X \backslash H}}$
    \State Let $\overline{\vec{v}} =  \Call{findMaxCorner}{
        \underline{\vec{v}},\Phi_{X \backslash H
        \backslash
        \{ \vec{v}~|~\exists \vec{v}'\in V: \underline{\vec{v}} \leq \vec{v}' \leq \vec{v}  \}
    }}$
    \State $V \gets V\uplus \{\underline{\vec{v}}\}$
    \State \textbf{return} $H \cup \Cube{\underline{\vec{v}}}{\overline{\vec{v}}}$
  \Else
    \State Let $\underline{\vec{v}} = \Call{findMinCorner}{\vec{v}_e,\Phi_{H\backslash X}}$
    \State Let $\overline{\vec{v}} =  \Call{findMaxCorner}{\underline{\vec{v}},
        \Phi_{H\backslash X
        \backslash
        \{ \vec{v}~|~\exists \vec{v}'\in V: \underline{\vec{v}} \leq \vec{v}' \leq \vec{v}  \}
    }}$
    \State $V \gets V\uplus \{\underline{\vec{v}}\}$
    \State \textbf{return} $H \backslash \Cube{\underline{\vec{v}}}{\overline{\vec{v}}}$
  \EndIf
  \EndFunction
\end{algorithmic}
\caption{Optimized refinement avoiding visited minimal corners}
\label{alg:opt}
\end{algorithm}

Notice how only the maximal corner search benefits from the optimization, by
tracking down minimal corners only. As a matter of fact, one could store the
whole visited cubes in set $V$. However, when a search for maximal corner
is carried, the resulting cube will intersect a previously visited cube as soon
as the max corner crosses the minimal corner of the visited cube.

We exploit again \cref{rem:oraclecomp} to build an oracle for every mentioned
membership oracle. Since $V$ is a finite set, one can indeed build a
membership oracle for the set
$\{\vec{v}~|~\exists \vec{v}'\in V\backslash X: \underline{\vec{v}} \leq \vec{v}' \leq \vec{v}\}$.
Due to this exclusion region, a
finer analysis has to be conducted to prove $H\in\intb$.

\begin{lemma}
  \label{lem:inv}
   The two optimized variants maintain the following invariants:
   \begin{enumerate}
     \item $V\cap X \subseteq H$;
     \item $(V\backslash X) \cap H = \emptyset$;
     \item for all $\vec{v}\in V$, and any $k$, $\vec{v}[k]\in \underline{B}_k$;
     \item $H\in \intb$.
   \end{enumerate}
\end{lemma}

Properties $1$ and $2$ ensure that every $v$ added to $V$ is never added twice.
These also ensures correctness of the algorithm: remark that the search for a
maximal corner is not started from the initial counterexample $\vec{v}_e$ but
from $\underline{\vec{v}}$, which is indeed is in the search space since
$\underline{\vec{v}} \notin
\{ \vec{v}~|~\exists \vec{v}'\in V: \underline{\vec{v}} \leq \vec{v}' \leq \vec{v}\}$
(no point added twice to $V$).
Finally, property $3$ ensures that only elements of $(\underline{B}_k)_k$ are
added to $V$, hence a maximal number of $(2n)^d$ additions.

\begin{proof}
  At the beginning of the algorithm, $V=H=\emptyset$, satisfying all given
  properties. We prove the result by induction on the iteration step:
  \begin{enumerate}
    \item By definition of corner oracles, namely
        \Call{FindMaxCorner}{}, if $\underline{\vec{v}}\in X$ has
        been added to $V$ during some previous iteration,
        it was added in the first branch
        (the oracle returns some point in the search region, which excludes
        $X$ in the second branch).
        Therefore, it was also added to $H$ during this iteration.
        Consider some later iteration removing elements from $H$, namely
        an iteration executing the second branch.
        Some cube
        $C=\Cube{\underline{\vec{v}}'}{\overline{\vec{v}}'}$ has
        been computed by the corner oracles in this branch such that
        $\overline{\vec{v}}' \in H\backslash X\backslash
            \{ v~|~\exists \vec{v}'\in V: \underline{\vec{v}}' \leq \vec{v}' \leq \vec{v}  \}$
        In particular, since $\underline{\vec{v}}\in V$, we do not have
        $\underline{\vec{v}}' \leq \underline{\vec{v}} \leq \overline{\vec{v}}'$ hence
        $\underline{\vec{v}}\notin C$ and $\underline{\vec{v}}$ is not removed.
     \item Similar to $(1)$ (symmetric case).
     \item For every $\underline{\vec{v}}$ added to $V$, it was
        produced by a (max) corner query made on $X\backslash H$ or
        $H\backslash X$.
        Both of these sets are in $\intb$ since $H\in\intb$ by induction hypothesis.
     \item Let us prove that the cube $C=\Cube{\underline{\vec{v}}}{\overline{\vec{v}}}$ currently
        added or removed satisfies $C\in\intb$ (hence $H\cup C, H\backslash C\in \intb$
        which will conclude the induction).
        We already have proven that
        $\underline{\vec{v}}\in(\underline{B}_k)_k$.
        We prove now that $\overline{\vec{v}}\in (\overline{B}_k)_k$ which is searched
        over the restricted state space $B=A\backslash 
        \{ \vec{v}~|~\exists \vec{v}'\in V: \underline{\vec{v}} \leq \vec{v}' \leq \vec{v}  \}$
        for $A=X\backslash H\in\intb$ or $A=H\backslash X\intb$.

        For any $k\in\inter{1}{d}$,
        $\overline{\vec{v}}+\vec{e}_k \notin B$ so either:
        \begin{itemize}
            \item $\overline{\vec{v}}+\vec{e}_k \notin A\in\intb$ so $\overline{\vec{v}}[k] \in \overline{B}_k$;
            \item or $\overline{\vec{v}}+\vec{e}_k \in 
                \{ \vec{v}~|~\exists \vec{v}'\in V: \underline{\vec{v}} \leq \vec{v}' \leq \vec{v}  \}$ but
                since $\overline{\vec{v}}$ is not in the set, there exists $\vec{v}'\in V$
                such that $\overline{\vec{v}}[k]+1 = \vec{v}'[k]$. Since
                $\vec{v}'[k] \in \underline{B}_k$, we have $\overline{\vec{v}}[k] \in \overline{B}_k$.
        \end{itemize}
        This concludes the proof.
  \end{enumerate}
\end{proof}

By combining~\cref{prop:membership} and~\cref{lem:inv}, we summarize the
complexity of our overshooting algorithms for a particular
target $X=\cup_{i=1}^n \Cube{\underline{\vec{v}}_i}{\overline{\vec{v}}_i}
   \in \conceptCl_d$.
\begin{theorem}[M+EQ]
   \label{cor:m+eq}
   Both variants of \Call{LearnCubes}{} terminates in at most $(2n)^d$
   iterations, where an iteration requires:
   \begin{enumerate}
     \item One equivalence query;
     \item One corner query, or equivalently, a
        linear number $O(\size{X})$ of membership queries.
   \end{enumerate}
\end{theorem}

This algorithm terminates in polynomial time, for fixed $d$,
in any representation of target $X$.
In particular, the result holds in the worst-case where
the representation of $X$ as a finite union of cubes is minimal.
As a matter of fact the presented exponential bound in $d$ is tight: there
exists a target $X\in \calC$ and a pair of corner and equivalence oracles
such that both algorithms terminate in exponential time.
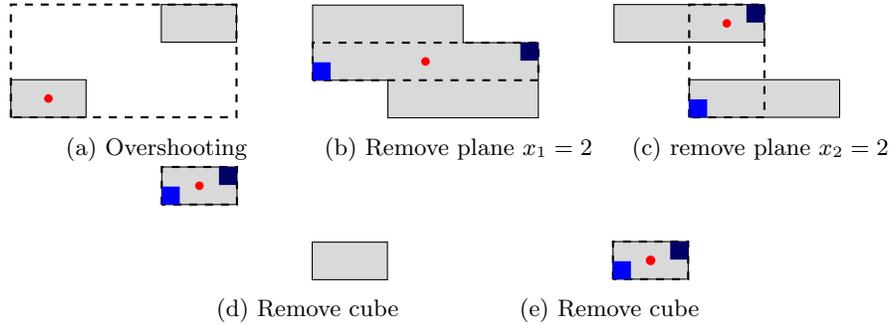
\begin{figure}
\centering
    \def\ax{0}\def\ay{0}
    \def\bx{1}\def\by{0.5}
    \def\cx{2}\def\cy{1.0}
    \def\dx{3}\def\dy{1.5}

    \def\shift{-3cm}
\begin{subfigure}[b]{0.32\textwidth}
  \begin{tikzpicture}
      \draw[box] (\ax,\ay) rectangle (\bx,\by);
      \draw[box] (\cx,\cy) rectangle (\dx,\dy);
      
      \node[cex] at ($(\ax,\ay)!0.5!(\bx,\by)$) {};
      \draw[learned] (\ax,\ay) rectangle (\dx,\dy);
  \end{tikzpicture}
  \caption{Overshooting}
\end{subfigure}
\begin{subfigure}[b]{0.32\textwidth}
  \begin{tikzpicture}
      \draw[box] (\ax,\by) -- (\ax,\dy) -- (\cx,\dy) -- (\cx,\cy) -- (\dx,\cy)
      -- (\dx,\ay) -- (\bx,\ay) -- (\bx,\by) -- (\ax,\by);
      
      \node[cex] at ($(\bx,\by)!0.5!(\cx,\cy)$) {};
      \draw[learned] (\ax,\by) node[mcorner] {} rectangle (\dx,\cy) node[Mcorner] {};
  \end{tikzpicture}
  \caption{Remove plane $x_1=2$}
\end{subfigure}
\begin{subfigure}[b]{0.32\textwidth}
  \begin{tikzpicture}
      \draw[box] (\ax,\cy) -- (\ax,\dy) -- (\cx,\dy) -- (\cx,\cy) -- (\ax,\cy);
      \draw[box] (\bx,\ay) -- (\bx,\by) -- (\dx,\by) -- (\dx,\ay) -- (\bx,\ay);
      
      \node[cex] at ($(\bx,\cy)!0.5!(\cx,\dy)$) {};
      \draw[learned] (\bx,\ay) node[mcorner] {} rectangle (\cx,\dy) node[Mcorner] {};
  \end{tikzpicture}
  \caption{remove plane $x_2=2$}
\end{subfigure}
\begin{subfigure}[b]{0.32\textwidth}
  \begin{tikzpicture}
      \draw[box] (\ax,\cy) -- (\ax,\dy) -- (\bx,\dy) -- (\bx,\cy) -- (\ax,\cy);
      \draw[box] (\cx,\ay) -- (\cx,\by) -- (\dx,\by) -- (\dx,\ay) -- (\cx,\ay);
      
      \node[cex] at ($(\ax,\cy)!0.5!(\bx,\dy)$) {};
      \draw[learned] (\ax,\cy) node[mcorner] {} rectangle (\bx,\dy) node[Mcorner] {};
  \end{tikzpicture}
  \caption{Remove cube}
\end{subfigure}
\begin{subfigure}[b]{0.32\textwidth}
  \begin{tikzpicture}
      \draw[box,draw=none,fill=none] (\ax,\cy) -- (\ax,\dy) -- (\bx,\dy) -- (\bx,\cy) -- (\ax,\cy);
      \draw[box] (\cx,\ay) -- (\cx,\by) -- (\dx,\by) -- (\dx,\ay) -- (\cx,\ay);
      
      \draw[learned] (\cx,\ay) node[mcorner] {} rectangle
        node[cex,midway] {}
        (\dx,\by) node[Mcorner] {};
  \end{tikzpicture}
  \caption{Remove cube}
\end{subfigure}
\caption{exponential blow-up, case $d=2$}
\label{fig:run2}
\end{figure}

\begin{example}
  \label{ex:expo}
Consider $X=\{\vec{0},\sum_{i=1}^d 2\vec{e}_i\}$ composed of two cubes, then
by learning $\Cube{\vec{0}}{\sum_{i=1}^d 2\vec{e}_i}$, then removing every middle
plane of equation $x_k=1$ for every $k\in\inter{1}{d}$, the resulting hypothesis
is composed of $2^d-2$ cubes to remove. An example with $d=2$ is depicted
in~\cref{fig:run2}.
\end{example}
Whether finite unions of cubes can be learned in polynomial time in the dimension
is left as an open problem, that we relate to DNF formula learning over
$d$~variables where each term can be interpreted as a cube over $\{0,1\}^d$.

\section{Extensions}
\label{sec:maximal}
In this section we introduce extensions to the overshooting algorithm from
\cref{ssec:overshooting}.
While membership and equivalence queries are sufficient for learning finite
sets, one natural extension of the minimal learner setting is to introduce
a subset oracle\cite{Ang88}:
\begin{definition}[Subset Oracle]
Consider some target concept $\concept\in \conceptCl_d$ for some concept class $\calC = \bigcup_{d=1}^\infty \conceptCl_d$
	and let $\bot,\top\notin \baseSet$ be two fresh symbols.
  \label{def:incloracle}
  
  A \emph{subset oracle} (\oracle{SUB}) for $\concept$ is a 
  function
  $\rho_\concept: \calC_d \rightarrow \{ \top, \bot \}$, which outputs 
  $\top$ iff $H \subseteq \concept$.
 
\end{definition}
The definition is similar to the membership oracle from \cref{def:meqoracles}
except the oracle takes a set instead of a single point as input.

\subsection{Maximal cube oracle}
\label{ssec:maximal}

\begin{figure}
  \begin{subfigure}[b]{0.5\textwidth}
  \centering
  \begin{tikzpicture}
    \def\ax{0}\def\ay{1}
    \def\bx{2}\def\by{2}
    \def\cx{1.5}\def\cy{1.5}
    \def\dx{3.5}\def\dy{2.5}

    \draw[box] (\ax,\ay) -- (\ax,\by) -- (\cx,\by) -- (\cx,\dy) -- (\dx,\dy) --
        (\dx,\cy) -- (\bx,\cy) -- (\bx,\ay) -- (\ax,\ay);
    \begin{scope}[xshift=-1, yshift=1]
      \draw[] (\ax,\ay) rectangle (\bx,\by);
      \draw[] (\cx,\cy) rectangle (\dx,\dy);
    \end{scope}
    \begin{scope}[xshift=1, yshift=-1]
      \draw[] (\ax,\cy) rectangle (\dx,\by);
      \draw[] (\cx,\ay) rectangle (\bx,\dy);
    \end{scope}
  \end{tikzpicture}
  \caption{$4$ maximal cubes when $n=2$}
  \label{sfig:fourmaxcube}
  \end{subfigure}
  \begin{subfigure}[b]{0.45\textwidth}
    \centering
    \begin{tikzpicture}
    \def\w{2}
    \def\h{0.2}
    \def\step{0.3}

    \draw[box] (0,0) rectangle (\w,\h);
    \draw[box] (\step,\h) rectangle +(\w,\h);
    \draw[box] ($2*(\step,\h)$) rectangle +(\w,\h);
    \node at ($3*(\step,\h)+0.5*(\w,\h)$) {$\hdots$};
    \draw[box] ($4*(\step,\h)$) rectangle +(\w,\h);
    \draw[box] ($5*(\step,\h)$) rectangle +(\w,\h);
    \end{tikzpicture}
    \caption{$n(n+1)/2$ maximal cubes}
    \label{sfig:quadmax}
  \end{subfigure}
  \caption{Example of maximal cubes w.r.t. to a union of $n$ cubes}
  \label{fig:maxcube}
\end{figure}
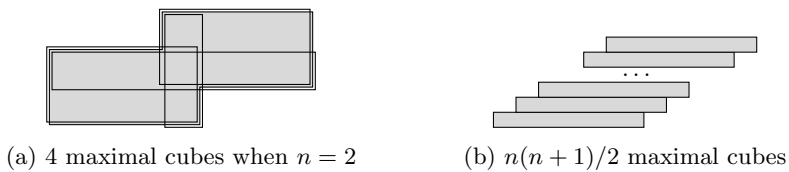

As opposed to the overshooting algorithm, using a subset oracle avoids the
overshooting issue, that is to say, we can now search for cubes included in the
target~$X$.
In order to increase the convergence speed, we nonetheless introduce a maximality
criterion on the suitable cubes:
\begin{definition}[Maximal Cubes]
  A cube $\Cube{\underline{\vec{v}}}{\overline{\vec{v}}}$ is \emph{maximal} w.r.t. $X$ if
  \begin{enumerate}
    \item $\Cube{\underline{\vec{v}}}{\overline{\vec{v}}} \subseteq X$
    \item For all i, $\Cube{\underline{\vec{v}}}{\overline{\vec{v}}+\vec{e}_i} \not\subseteq X$
    \item For all i, $\Cube{\underline{\vec{v}}-\vec{e}_i}{\overline{\vec{v}}} \not\subseteq X$
  \end{enumerate}
\end{definition}
\Cref{fig:maxcube} provides examples of possible maximal cubes in
dimension~$d=2$.

Next, we modify the corner oracle from \cref{ssec:corneroracle} to use subset
queries. 
Again, we only define the algorithm to find a max corner, the min corner
algorithm can be implemented analogously.
The algorithm first computes a lower and upper bound for the subsequent binary search.
The computation is shown in the function \Call{computeMaxBounds}{}.
Given a cube defined by its minimal and a maximal corner, the value of
coordinate~$i$
is increased as long as the resulting cube is still a subset of
the target set $X$.
The upper bound~$\overline{\vec{v}}$ is the first negative reply by the oracle and the lower bound~$\underline{\vec{v}}$
the last positive response.
A binary search is made on $\underline{\vec{v}}$ and ~$\overline{\vec{v}}$ in the \Call{findMaxIncCorner}{} function.

\begin{algorithm}[h]
\begin{algorithmic}
	\Ensure Returned value is a maximal corner of~$X$
	\Function{findMaxIncCorner}{${\underline{\vec{v}}}, {\overline{\vec{v}}}, \rho_X$}
	\For{$i\in \inter{1}{d}$}
	    \State $({\underline{b}}, {\overline{b}}) = \Call{computeMaxBounds}{{\underline{\vec{v}}},{\overline{\vec{v}}}, i, \rho_X}$
	    \While{${\underline{b}} \neq {\overline{b}}$}
	        \State{$m \gets ({\underline{b}} + {\overline{b}}) \div 2$}
	        \If{$\rho_X(\Cube{{\underline{\vec{v}}}}{{\overline{\vec{v}}[i/m]}})$}
	            \State ${\underline{b}} \gets m$
	        \Else
	            \State ${\overline{b}} \gets m$
	        \EndIf
	    \EndWhile
        \State $\overline{\vec{v}}[i] \gets \overline{b}$
	\EndFor
	\State \Return ${\overline{\vec{v}}}$
	\EndFunction
	\Function{computeMaxBounds}{$\underline{\vec{v}}, \overline{\vec{v}}, i, \rho_X$}
	\State{${\delta} \gets 1$}
	\While{$\rho_X(\Cube{{\underline{\vec{v}}}}{{\overline{\vec{v}}+{\delta}\cdot \vec{e}_i}})$}
	\State{${\delta} \gets {2\cdot \delta}$}
	\EndWhile
    \State \Return $(\overline{\vec{v}}[i]+\delta/2,\overline{\vec{v}}[i]+{\delta})$
	\EndFunction
\end{algorithmic}
\caption{Maximal corner of a maximal cube, in $O(\size{X})$ subset queries}
\label{alg:findmaxinc}
\end{algorithm}

\subsection{Maximal cube algorithm}
\Cref{alg:maxcube} presents a procedure that iteratively refines the hypothesis:
for any point, the algorithm searches for a maximal cube contained by this point
w.r.t. the target and adds it to the hypothesis.
One can check that both procedure calls are valid, as $H \subseteq X$ is an
invariant. At every iteration the counterexample $\vec{v}$ satisfies $\vec{v} \in X \setminus H$.
The use of the subset oracle ensures that the function \Call{FindMaxIncCorner}{} always
returns a point $\overline{\vec{v}}$ such that $\Cube{\vec{v}}{\overline{\vec{v}}}\subseteq X$.
Similarly,  the function \Call{FindMinIncCorner}{} always returns a corner
$\underline{\vec{v}}$ such that $\Cube{\underline{\vec{v}}}{\overline{\vec{v}}}\subseteq X$.
The resulting cube is then added to the hypothesis, ensuring point $\vec{v}$ is
never visited again as a counterexample. This entails the termination
of the algorithm, in at most $|X|$ iteration of the main loop. A better bound
will be explored in~\cref{ssec:complexity}.
\begin{algorithm}[h]
	\begin{algorithmic}
		\Function{LearnMaxCube}{$\rho_X$, $\Psi_X$}
		\State Let $H\gets \emptyset$
		\While{$(\vec{v} \gets \Psi_X(H))\neq \top$}
		\State Let $\overline{\vec{v}} \gets \Call{findMaxIncCorner}{\vec{v}, \vec{v},
			\rho_X
		}$
		\State Let $\underline{\vec{v}} \gets \Call{findMinIncCorner}{\vec{v}, \overline{\vec{v}},
			\rho_X
		}$
		\State $H\gets H\cup \Cube{\underline{\vec{v}}}{\overline{\vec{v}}}$
		\EndWhile
		\EndFunction
	\end{algorithmic}

	\caption{The maximal cube algorithm}
    \label{alg:maxcube}
\end{algorithm}

\subsection{Extension to the infinite case}
\label{sec:infinity}
We discuss now one possible extension to the infinite case, namely
when cubes are possibly unbounded and may contain
infinitely many points.

We adapt our learning formalism to deal with infinite bounds:
for the remainder of the section we extend the discrete lattice $\mathbb{Z}^d$ to
$\bbZi^d$ and extend trivially $\leq$ over the newly introduced points.
For $\underline{\vec{v}},\overline{\vec{v}}\in\bbZi^d$,
the definition of $C=\Cube{\underline{\vec{v}}}{\overline{\vec{v}}}$ remains unchanged, in
particular $C\subseteq \bbZ^d$ but may be infinite. The concept class $\calC$,
hence the domain of oracle functions,
is augmented with all finite unions of cubes with (possibly) infinite bounds.

A possible approach to tackle this problem in the minimally adequate teacher (M+EQ)
formalism consists in running the overshooting algorithm of~\cref{sec:exact}
on the state space restricted to some cube of width $2^k$ centered in $\vec{0}$
and gradually increase $k$ if counterexamples outside this restriction are found.
\ifextended
This method is discussed in \cref{sec:infinitymemeq}
\else
This method is discussed in the extended version of the present article~\cite{techreport}
\fi but we focus here on a
\Call{LearnMaxCube}{} adaptation exploiting subset queries (SUB+EQ).

While \cref{alg:maxcube} remains unchanged, we need however to adjust the
functions \Call{FindMaxIncCorner}{} and
\Call{FindMinIncCorner}{} as those are not able to accelerate the search to infinity.
\Cref{fig:maxbounds} achieves this goal by simply
overriding the \Call{ComputeMaxBounds}{} and \Call{ComputeMinBounds}{}
subroutines in order to check for possible $+\infty$ and $-\infty$ bounds.
Whenever such bound is returned, no further binary search occurs for this
coordinate (constant time).

\begin{algorithm}
	\begin{algorithmic}
	\Function{computeMaxBounds}{${\underline{\vec{v}}}, {\overline{\vec{v}}}, i, \rho_X$}
	  \If{$\rho_X(\Cube{{\underline{\vec{v}}}}{{\overline{\vec{v}}[i/+\infty]}})$}
	    \State \Return $(+\infty,+\infty)$
	  \Else \Comment{We refer to original \Call{ComputeMaxBounds}{} of~\cref{alg:findmaxinc}}
	    \State \Return \Call{Super.ComputeMaxBounds}{$\underline{\vec{v}},\overline{\vec{v}},i,\rho_X$}
	  \EndIf
	\EndFunction
	\end{algorithmic}
\caption{Maximal bound overriding, checking for $+\infty$.}
\label{fig:maxbounds}
\end{algorithm}

\subsection{Complexity}
\label{ssec:complexity}
Termination of \Call{LearnMaxCube}{} was proved using cardinality arguments
in~\cref{ssec:maximal}. These arguments obviously don't apply in the case
where the target set is infinite. Moreover, we are interested in finer
complexity analysis.

As in~\cref{ssec:compovershoot}, we fix a target representation
$X=\cup_{i=1}^n \Cube{\underline{\vec{v}}_i}{\overline{\vec{v}}_i}$
and
study the algorithm complexity with respect to
$\sum_{i=1}^n \size{\underline{\vec{v}}_i} + \size{\overline{\vec{v}}_i}\in
\conceptCl_d$.
As some of the vectors $\vec{v}$ may contain infinite coordinates, we carefully specify
$\size{+\infty}=\size{-\infty}=1$ and keep the usual definition of $\size{v}$.

\begin{theorem}[SUB+EQ]
   \label{thm:maxcube}
   \Call{LearnMaxCube}{} terminates in at most $n^{2d}$
   iterations, where an iteration requires:
   \begin{enumerate}
     \item One equivalence query;
     \item One maximal cube query, or equivalently, a
        linear number $O(\size{X})$ of subset queries.
   \end{enumerate}
\end{theorem}
\begin{proof}
    At every iteration,
    one equivalence query is performed then
    \Call{FindMaxIncCorner}{} and \Call{FindMinIncCorner}{}
    perform a binary search, resulting in a linear number of subset
    similar (proof similar to \cref{prop:membership}).

    In order to analyze the number of iterations of the main loop, let us
    first remark that each added maximal cube is added only once: 
    if we write $\vec{v}_k$ the $k$-th counterexample and $C_k$ the learned maximal cube,
    then $\vec{v}_{k+1}\in X\backslash \cup_{i=1}^{k} C_i$ and $\vec{v}_{k+1}\in C_{k+1}$
    so $C_{k+1} \neq C_i$ for every $i\in \inter{1}{k}$.

    The number of iterations is therefore bounded by the number of maximal
    cubes. We proceed now to bound the number of maximal cubes:
    Let $C=\Cube{\underline{\vec{v}}}{\overline{\vec{v}}}$ be a maximal cube w.r.t.~$X$.
    For any $k\in\inter{1}{d}$ there exist $i,j\in\inter{1}{n}$ such that
    $\underline{\vec{v}}[k] = \underline{\vec{v}}_i[k]$ and
    $\overline{\vec{v}}[k] = \overline{\vec{v}}_j[k]$, hence at most $n^2$ possibilities
    for coordinate~$k$.
\end{proof}

As in \cref{cor:m+eq} the number of iterations is polynomial in the number of
cubes~$n$ but exponential in the dimension~$d$. As opposed to the \Call{LearnCubes}{}
algorithm, the bound is not tight as the example \cref{sfig:quadmax}
provides only a quadratic number of maximal number of cubes.
As the maximal cube concept can be related to the notion of \emph{prime
implicant}, examples of DNF formula with an exponential of prime implicants
(see for example~\cite{chandra87}) can be translated into union of cubes with
an exponential number of maximal 0-1 cubes.

From a practical perspective, one can nonetheless argue that
\Call{LearnMaxCube}{} is likely to perform well in practice, by avoiding the
overshooting problem mentioned in \cref{ex:expo} as $H\subseteq X$ is an invariant.
In fact, one can easily check that if there are no adjacent\footnote{
two cubes $C_1$ and $C_2$ are \emph{adjacent} if
$\min\left\{\sum_{i} |\vec{v}_1[i] - \vec{v}_2[i]\middle| ~|~ \vec{v}_1\in C_1,\vec{v}_2\in C_2\right\} \leq 1$}
cubes, the number of iterations becomes linear.

\section{Applications and Experiments}
\label{sec:exp}

In this section, we describe an immediate application of our learning 
algorithms to monadic decomposition of quantifier-free Presburger formulas
\cite{monadic-decomposition,HLRW20}. We then report on experimental comparisons
between our algorithms and existing methods for the problem.

\subsection{Application to Monadic Decomposition}
Here we consider quantifier-free linear integer arithmetic formulas without 
modulo arithmetic:
\[
    \varphi ::= \alpha_1 \sim \alpha_2\ |\ \varphi \wedge \varphi\ |\ \varphi \vee
                                    \varphi,
\]
where $\sim\ \in \{\leq,\geq,=\}$, and $\alpha_1, \alpha_2$ are integer linear
combinations of the variables $x_1,\ldots,x_n$, i.e.,
$\alpha_i$ is of the form $c_0 + \sum_{j=1}^n c_j.x_j$, where each
$c_i \in \Z$. The formula $\varphi(\bar x)$ is said to be \emph{satisfiable}
(written $\IntTh \models \varphi$) if there
exists an assignment $\sigma$ of $\bar x$ to $\Z$ such that the formula becomes
true. Of course, this is just a simple fragment of the first-order theory of 
integer linear arithmetic and the notion of $\IntTh \models \varphi$ can be
defined in the same way even with quantifiers \cite{KS08,survival}. A formula 
$\varphi$ is said to be \emph{monadic} if it has only
one variable. Every monadic formula $\varphi(x)$ in this fragment can be easily 
transformed into a union integer intervals of the form: (1) $l \leq x \wedge
x \leq u$ where $l,u \in \Z$, (2) $l \leq x$ where $l \in \Z$, (3) $x \leq 
u$ where $u \in Z$, or (4) $\top$ or $\bot$. 

A \emph{monadic decomposition} \cite{monadic-decomposition} of a formula 
$\varphi(\bar x)$ is a boolean 
combination $\psi(\bar x)$ of monadic formulas that is equivalent to $\varphi$
over the theory, i.e., 
$\IntTh \models \forall \bar x( \varphi \leftrightarrow \psi )$.
Of course, not all formulas admit a monadic decomposition (e.g., $x = y$).
It was shown in \cite{HLRW20} that deciding if a formula in the theory be 
monadically decomposable is coNP-complete\footnote{The proof in \cite{HLRW20}
uses modulo constraints to show that monadic decomposition of a two-variable
formula $\varphi(x,y)$ is coNP-complete. Modulo constraints could be easily
removed by allowing more integer variables.}. Veanes \emph{et al.}
\cite{monadic-decomposition} provides a generic semi-decision procedure for 
computing a monadic decomposition of a quantifier-free formula as an
if-then-else formula that is
applicable to pretty much all theories considered in SMT. Despite the
genericity, the procedure runs rather well, e.g., as the authors showed on
their benchmarking in \cite{monadic-decomposition}. 

The application of our learning algorithms to computing monadic decomposition
arises from the following observation.
Since each monadic decomposition can be transformed into DNF, a monadic 
decomposition of a formula $\varphi(\bar x)$ over $\IntTh$ can be constructed as 
a finite 
union of (possibly infinite) hypercubes, where an infinite hypercube arises when
a variable is either not bounded from above or not bounded from below (or both).
Conversely, a finite union $H$ of possibly infinite hypercubes can also be 
easily transformed into a boolean combination of monadic formulas $\varphi_H$.
For example, the formula
$(0 \leq x \leq 5 \land 3 \leq y \leq 10) \vee (8 \leq x)$ corresponds to the
union of hypercubes $\Cube{(0,3)}{(5,10)} \cup \;\Cube{(8,-\infty)}{(+\infty,+\infty)}$.
Furthermore, all relevant oracles admit a straightforward implementation:
\begin{itemize}
\item A membership query $\bar v$ requires checking $\IntTh \models 
\varphi(\bar v)$, which can be checked in polynomial-time because $\varphi$ is
quantifier-free.
\item An equivalence query $H$ can be reduced to checking
    \[
        \IntTh \models (\varphi_H \wedge \neg \varphi) \vee (\varphi \wedge \neg
            \varphi_H).
    \] 
    This is a single satisfiability check of quantifier-free integer-linear 
        arithmetic formula, for which highly-optimized solvers exist (e.g., Z3 
        \cite{z3}).
\item A subset query $H$ can similarly be reduced to checking
    \[
        \IntTh \models (\varphi_H \wedge \neg \varphi).
    \] 
    This is also a single satisfiability check over $\IntTh$.
\end{itemize}
This allows us to apply both of our learning algorithms to the problem.

{\color{black}
Monadic decomposition has numerous applications including 
quantifier elimination
\cite{monadic-decomposition}, string solving \cite{HLRW20}, and symbolic finite
automata/transducers \cite{monadic-decomposition,symbolic-transducer-power},
among others. In the following example we illustrate how our learning 
algorithm(s) could be applied to improving quantifier elimination for the 
theory of linear integer arithmetic. 
\begin{example} \label{ex-mondec}
    Consider a formula of the form $\forall \bar x\exists y~\varphi(\bar x,\bar
    y)$, where $\varphi$ is a formula in linear integer arithmetic without
    modulo constraints. Suppose that $\varphi$ is monadically decomposable, and is
    equivalent to the formula $\bigvee_{i=1}^n D_i(\bar x,\bar y)$, where each 

    $D_i$ is a disjunction of monadic predicates over the variables $\bar x 
    \cup \bar y$. We assume w.l.o.g. that each $D_i$ is satisfiable. 
    Then, this formula is equisatisfiable (over linear integer
    arithmetic) to
        $\psi := \forall \bar x\left( \bigvee_{i=1}^n D_i(\bar x,\bar
        c_i)\right),$
    where $\bar y$ in $D_i$ are replaced by \emph{fresh} constants $\bar c_i$
    (i.e. two distinct $D_i,D_i'$ use different constants). This can be proven
    by a simple application of skolemization, and observing that each
    occurrence of $f(\bar x)$ in any disjunct is of the form $a < f(\bar x) <
    b$, where $a \in \{-\infty\} \cup \mathbb{Z}$ and $b \in \mathbb{Z} \cup
    \{\infty\}$, implying that $f(\bar x)$ can be replaced by a single constant,
    which does not depend on $\bar x$. Finally, let
    $D_i'$ be the conjuncts in $D_i$ only involving variables in $\bar x$.
    Checking that $\psi$ is true reduces to checking satisfiability of 
    $\bigwedge_{i=1}^n \neg D_i'$.

    To make this example concrete, we consider the formula
    $\forall x \exists y (x \geq 0 \rightarrow x + y \geq 5 \wedge y \geq 0)$.
    A monadic decomposition of the quantifier-free part is
    $
        x < 0 \vee \bigvee_{i=0}^5 (x\geq i \wedge y \geq 5-i)$.
    Therefore, checking the above formula can be reduced to satisfiability of
        $x \geq 0 \wedge \bigwedge_{i=0}^5 x < i$
    which is not satisfiable.
    \OMIT{
    whose quantifier-free is a simplification of 
    which is a simplification of 
    which incidentally corresponds to the 1 Diagonal Restricted benchmark
    below with extra variable quantifications. The quantifier-free part admits
    the monadic decomposition 
    $x \leq 1 \land y \leq 1 \land x \geq 1 \land y \geq 1$

Here these techniques result in the equivalent formula
$\forall x (x \leq 3 \land f(x) \leq 3 \land x \geq 1 \land f(x) \geq 1 \land x +f(x) \leq 2)$.
The introduction of the unary Skolem function $f$ can be avoided in this case by
computing the monadic decomposition $x \leq 1 \land y \leq 1 \land x \geq 1 \land y \geq 1$
where $y$ can be replaced by
an uninterpreted constant.
    }
\end{example}
}



\subsection{Experiments}
{\color{black}
In order to assess the performance of the algorithms
\Call{FindMaxCorner}{} and \Call{FindMinCorner}{}
respectively introduced in
\cref{sec:exact} and \cref{sec:maximal}, we consider prototype implementations.
The following prototypes and experiments can be found in~\cite{artifact}.

\paragraph{Variants.}
Although the methods were presented with binary search strategies in mind,
we also implemented a more naive unary search procedure to obtain the corners.
As later noticed in the experiments, unary search may be preferred for very small
cubes and performs especially well for cubes which are based 0-1 integer programs,
while binary search achieves better performance for larger cubes.
Consequently,
we refer to a third variation of the
algorithm called \enquote{optimized}, combining unary search for small instances
and binary search for large values.
More precisely two variants of the overshooting algorithm from \cref{sec:exact}
and three variants of the max cubes algorithm from \cref{sec:maximal} are
presented, called respectively
\emph{overshoot\_unary} and \emph{overshoot\_binary} and
\emph{max\_unary}, \emph{max\_binary} and \emph{max\_optimized}.

\paragraph{Tool comparison.}
Evaluation is performed against a generic monadic decomposition procedure
\emph{mondec$_1$} from \cite{monadic-decomposition} by Veanes et al, which
works over an arbitrary base theory and outputs an if-then-else formula, which
could be exponentially more succinct than a formula in DNF.
The algorithm, which exploits the python-Z3 framework~\cite{z3}, uses a kind of 
a decision tree search heuristics to split the input into monadic predicates.

\paragraph{Implementation.}
Similarly to mondec$_1$, our prototype is implemented in python using the
python-Z3 framework, but is specialized in handling linear integer
arithmetic formula, and that outputted formulas will be in DNF, unlike
mondec$_1$.
For monadic decomposition applications, oracles queries are converted to 
appropriate Z3 satisfaction queries since 
a (possibly non-monadic) representation of the target set is 
already known.
\OMIT{
Moreover,
direct oracle implementations remain relevant in the monadic decomposition
setting where a (possibly non-monadic) representation of the target set is 
already known.
}

\begin{figure}
	\begin{center}
		\begin{subfigure}[t]{0.55\linewidth}
			\includegraphics[scale=0.3]{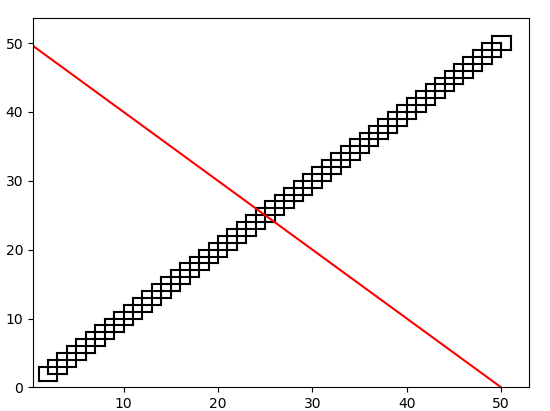}
			\caption{50 overlapping cubes and the diagonal \mbox{$x+y = 50$}.}\label{fig:o-1}
		\end{subfigure}
		\begin{subfigure}[t]{0.35\linewidth}
			\includegraphics[scale=0.3]{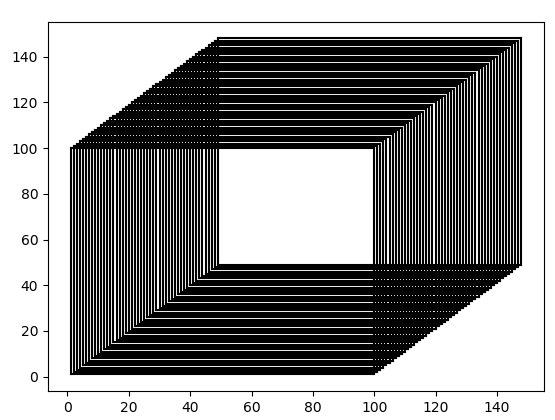}
			\caption{100 big Cubes.}\label{fig:o-4}
		\end{subfigure}
	\end{center}
	\caption{Benchmarks for $\mathbb{Z}^2$.}
\end{figure}
\subsection{Benchmark suite}
%
%
%

Our benchmark suite is restricted to the problem of monadic decomposition of
linear integer arithmetic, and its purpose is to stress-test our learning 
algorithms and mondec$_1$ against various kinds of ``extreme conditions''.
The suite consists of six classes of monadically decomposable example formulas,
which were constructed to test five features (see below). 
Note that the given formulas themselves might contain non-monadic predicates.
\begin{wraptable}{r}{7.5cm}
	\centering
	\resizebox{0.6\columnwidth}{!}{
		\begin{tabular}{cccccc}
			\hline
			& Overlap & \# Cubes & $|$Cube$|$ & Dimension & Unbounded   \\
			\hline
            (a)
            & + & + & - & - & -   \\
			
            (b)
            & + & - & - & + &  -  \\
			
            (c)
            & + & + & - & - & -  \\
			
            (d)
            & + & + & + & - & -  \\
			
            (e)
            & - & + & - & - & -   \\
			
            (f)
            & + & + & + & - & +   \\
			
	\end{tabular}}
	\caption{Features of conducted benchmarks.
        A ``+'' (resp. ``-'') indicates a high (resp.~low) presence of a 
        feature.
    }
    \label{tab:features}
\end{wraptable}
The five features (left to right in Table \ref{tab:features}) represent the 
presence of (1) a large amount of cube overlaps, (2) 
a large number of cubes, (3) a large cube, (4) large dimension, and (5)
an unbounded cube.
We hypothesized that these five features play important roles in how fast the
algorithms perform, which are indeed validated in our experimental results.
The six classes of formulas are elaborated below.

\begin{itemize}
  \item[(a)] 
    \emph{K Diagonal Restricted} consists of K overlapping cubes of length and
    width 2 and one diagonal as shown in \cref{fig:o-1}. The cubes overlap with
    at most two other cubes and stack up diagonally. The algorithms need to
    return all the cubes left of the diagonal.
  \item[(b)]
    \emph{10 cubes in $\mathbb{Z}^d$} consists of $K=10$ overlapping
    cubes of size $2^d$ stacking up diagonally similar to the benchmark K Diagonal
    Restricted without diagonal restriction.
  \item[(c)]
    \emph{K Diagonal Unrestricted} is a variation of~\cref{fig:o-1} where
    the algorithms
    need to return \emph{all} the cubes and all the points on the diagonal.
  \item[(d)] 
    \emph{K Big Overlapping Cube} is a benchmark testing large cubes as depicted
    in \cref{fig:o-4}.
    It consists of K overlapping cubes of length and width 100 and
    are overlapping and stacking up diagonally like the benchmark K Diagonal
    Restricted.
  \item[(e)] 
    \emph{K Diagonal} is built as the set of points along the
    diagonal $x = y \leq K$.
  \item[(f)]
    \emph{Example~\ref{ex-mondec}} is generalized to any $K\in\mathbb{N}$ by
    $x \geq 0 \rightarrow x + y \geq K \wedge y \geq 0$. Its unbounded nature
    makes it tractable by max\_optimized and mondec$_1$ only.
\end{itemize}

\subsection{Results}
Experiments were conducted on an AMD Ryzen 5 1600 Six-Core CPU with 16 GB of RAM running on Windows 10.
The results are summarized in \cref{tab:benchmarks} where each graph represents
one benchmark comparing the run times of each algorithm.


\begin{figure}
    \centering
    \def\scale{1}
    \def\scalesub{0.45}
    \def\gridwidth{6.3cm}
    \def\gridheight{4.5cm}
    \def\colspace{0.01\textwidth}
	\begin{subfigure}[t]{\scalesub\linewidth}
		\pgfplotstableread{diagonal-restricted.dat}{\table}
		
		\begin{tikzpicture}[scale=\scale]
			\begin{axis}[
				xmin = 50, xmax = 500,
				ymin = 0, ymax = 1800,
				xtick distance = 50,
				ytick distance = 200,
				grid = none,
				minor tick num = 5,
				major grid style = {lightgray},
				minor grid style = {lightgray!25},
                width = \gridwidth,
                height = \gridheight,
				legend cell align = {left},
				legend pos = south east,
				legend style={
					at={(0,0)},
					anchor=north,at={(axis description cs:0.3,-0.1)}}
				]
				
				\addplot[blue] table [x = {x}, y = {y1}] {\table};
				
				\addplot[red] table [x ={x}, y = {y2}] {\table};
				
				\addplot[teal] table [x = {x}, y = {y3}] {\table};
				
				\addplot[purple] table [x = {x}, y = {y4}] {\table};
				\addplot[orange] table [x = {x}, y = {y5}] {\table};
				\addplot[gray] table [x = {x}, y = {y6}] {\table};
				
			\end{axis}
		\end{tikzpicture}
		\caption{Benchmark on K Diagonal Restricted in $\mathbb{Z}^2$.\\ The x-axis encodes the amount of cubes $K$.}
		\label{fig:k-dia-res}
	\end{subfigure}
    \hspace{\colspace}
	\begin{subfigure}[t]{\scalesub\linewidth}
		\pgfplotstableread{dimension-d.dat}{\table}
		
		\begin{tikzpicture}[scale=\scale]
			\begin{axis}[		
				xmin = 3, xmax = 200,
				ymin = 0, ymax = 1800,
				xtick distance = 50,
				ytick distance = 200,
				grid = none,
				minor tick num = 5,
				major grid style = {lightgray},
				minor grid style = {lightgray!25},
				width = \gridwidth,
				height = \gridheight,
				legend cell align = {left},
				legend pos = south east,
				legend style={
					at={(0,0)},
					anchor=north,at={(axis description cs:0.3,-0.1)}}
				]
				
				\addplot[blue] table [x = {x}, y = {y1}] {\table};
				
				\addplot[red] table [x ={x}, y = {y2}] {\table};
				
				\addplot[teal] table [x = {x}, y = {y3}] {\table};
				
				\addplot[purple] table [x = {x}, y = {y4}] {\table};
				\addplot[orange] table [x = {x}, y = {y5}] {\table};
				\addplot[gray] table [x = {x}, y = {y6}] {\table};
			\end{axis}
			
		\end{tikzpicture}
		\caption{Benchmark on 10 cubes in $\mathbb{Z}^d$.\\
        The x-axis encodes the dimension $d$.}
		\label{fig:dimension-d}
    \vspace{1.3cm}
	\end{subfigure}
	\begin{subfigure}[t]{\scalesub\linewidth}
		\pgfplotstableread{diagonal-unrestricted.dat}{\table}
		
		\begin{tikzpicture}[scale=\scale]
			\begin{axis}[		
				xmin = 50, xmax = 500,
				ymin = 0, ymax = 1800,
				xtick distance = 50,
				ytick distance = 200,
				grid = none,
				minor tick num = 5,
				major grid style = {lightgray},
				minor grid style = {lightgray!25},
			    width = \gridwidth,
			    height = \gridheight,
				legend cell align = {left},
				legend pos = south east,
				legend style={
					at={(0,0)},
					anchor=north,at={(axis description cs:0.3,-0.1)}}
				]
				
				\addplot[blue] table [x = {x}, y = {y1}] {\table};
				
				\addplot[red] table [x ={x}, y = {y2}] {\table};
				
				\addplot[teal] table [x = {x}, y = {y3}] {\table};
				
				\addplot[purple] table [x = {x}, y = {y4}] {\table};
				\addplot[orange] table [x = {x}, y = {y5}] {\table};
				\addplot[gray] table [x = {x}, y = {y6}] {\table};
				
			\end{axis}
			
		\end{tikzpicture}
		\caption{Benchmark on K Diagonal Unrestricted in $\mathbb{Z}^2$.\\
        The x-axis encodes the amount of cubes $K$.}
		\label{fig:k-dia-unre}
	\end{subfigure}
    \hspace{\colspace}
	\begin{subfigure}[t]{\scalesub\linewidth}
		\pgfplotstableread{big-cubes.dat}{\table}
		
		\begin{tikzpicture}[scale=\scale]
			\begin{axis}[		
				xmin = 50, xmax = 500,
				ymin = 0, ymax = 1800,
				xtick distance = 50,
				ytick distance = 200,
				grid = none,
				minor tick num = 5,
				major grid style = {lightgray},
				minor grid style = {lightgray!25},
				width = \gridwidth,
				height = \gridheight,
				legend cell align = {left},
				legend pos = south east,
				legend style={
					at={(0,0)},
					anchor=north,at={(axis description cs:0.3,-0.1)}}
				]
				
				\addplot[blue] table [x = {x}, y = {y1}] {\table};
				
				\addplot[red] table [x ={x}, y = {y2}] {\table};
				
				\addplot[teal] table [x = {x}, y = {y3}] {\table};
				
				\addplot[purple] table [x = {x}, y = {y4}] {\table};
				\addplot[orange] table [x = {x}, y = {y5}] {\table};
				\addplot[gray] table [x = {x}, y = {y6}] {\table};
				
			\end{axis}
			
		\end{tikzpicture}
		\caption{Benchmark on K Big Cubes in $\mathbb{Z}^2$.\\ The x-axis encodes the amount of cubes $K$.}
		\label{fig:k-big-cubes}
        \vspace{0.8cm}
	\end{subfigure}

\begin{subfigure}[t]{\scalesub\linewidth}
	\pgfplotstableread{diagonal-single-points.dat}{\table}
	
	\begin{tikzpicture}[scale=\scale]
		\begin{axis}[		
			xmin = 50, xmax = 500,
			ymin = 0, ymax = 250,
			xtick distance = 50,
			ytick distance = 200,
			grid = none,
			minor tick num = 5,
			major grid style = {lightgray},
			minor grid style = {lightgray!25},
			width = \gridwidth,
			height = \gridheight,
			legend cell align = {left},
			legend pos = south east,
			legend style={
				at={(0,0)},
				anchor=north,at={(axis description cs:0.3,-0.1)}}
			]

            \addplot[blue] table [x = {x}, y = {y5}] {\table};
            \addplot[red] table [x ={x}, y = {y6}] {\table};
            \addplot[teal] table [x = {x}, y = {y3}] {\table};
            \addplot[purple] table [x = {x}, y = {y4}] {\table};
            \addplot[orange] table [x = {x}, y = {y1}] {\table};
            \addplot[gray] table [x = {x}, y = {y2}] {\table};

		\end{axis}
		
	\end{tikzpicture}
	\caption{Benchmark on K Diagonal in $\mathbb{Z}^2$.\\ The x-axis encodes the maximal value for x and y.}
	\label{fig:k-dia}
\end{subfigure}
\hspace{\colspace}
\begin{subfigure}[t]{\scalesub\linewidth}
	\pgfplotstableread{mondec-implies.dat}{\table}
	
	\begin{tikzpicture}[scale=\scale]
		\begin{axis}[		
			xmin = 50, xmax = 500,
			ymin = 0, ymax = 250,
			xtick distance = 50,
			ytick distance = 200,
			grid = none,
			minor tick num = 5,
			major grid style = {lightgray},
			minor grid style = {lightgray!25},
			width = \gridwidth,
			height = \gridheight,
			legend cell align = {left},
			legend pos = south east,
			legend style={
				at={(0,0)},
				anchor=north,at={(axis description cs:0.3,-0.1)}}
			]
			
			\addplot[orange] table [x = {x}, y = {y1}] {\table};
			
			\addplot[gray] table [x ={x}, y = {y2}] {\table};

		\end{axis}
		
	\end{tikzpicture}
	\caption{Benchmark on Example \ref{ex-mondec} in $\mathbb{Z}^2$.\\ The x-axis encodes parameter $K$.}
	\label{fig:mondec-implies}
\end{subfigure}

    \begin{center}
		\begin{tikzpicture}
            \begin{axis}[
                  hide axis,
    xmin=50,
    xmax=50,
    ymin=0,
    ymax=0, 
                legend columns=3,
                legend style={/tikz/every even column/.append style={column sep=0.5cm}}]
            ]
                \addplot[blue] {x};
                \addplot[red] {x};
                \addplot[teal] {x};
                \addplot[purple] {x};
                \addplot[orange] {x};
                \addplot[gray] {x};
				\legend{
					overshooting\_unary,
					overshooting\_binary,
					maxcube\_unary,
					maxcube\_binary,
					maxcube\_optimized,
					mondec$_1$
				}
            \end{axis}
		\end{tikzpicture}
        \vspace{-15em}
     \end{center}
	\caption{Benchmark results. The y-axis encodes the time in seconds. The timeout is set to 1800s.}
	\label{tab:benchmarks}
\end{figure}

{\color{black}
The overshooting phenomenon can be observed in~\cref{fig:k-dia-unre} and
\cref{fig:k-dia} with its quadratic shape, as $d=2$.
In \cref{fig:dimension-d}, the running time quickly diverges as $d$ increases,
as anticipated by \cref{ex:expo}.

When the considered cubes are small, as in \cref{fig:k-dia-res} and
\cref{fig:k-dia-unre}, the unary search algorithms outperform their binary
counterparts, meaning the few additional queries made by the binary search are
more costly than a direct enumeration. The optimized variant is therefore a good
compromise in all cases.

\Cref{fig:k-big-cubes} depicts a benchmark with many large cubes for a
fixed dimension. While the impact of the overshooting phenomenon remains
contained, the maxcube unary search variant is particularly slow. This can be
explained by the size of the cubes making unary search inefficient, combined
with the already expensive cost of every single inclusion query.

The mondec$_1$ algorithm is comparable to 
the overshooting algorithms in \cref{fig:k-dia}. It also
performs particularly well in \cref{fig:mondec-implies}, which we conjecture
is due to the conciseness of the solution in if-then-else form used by
mondec$_1$.

Overall, the maxcube algorithm in its optimized form is the most stable algorithm
for this benchmark set and should be preferred when an inclusion oracle is available.
The extra cost of these queries are here taken into account and remain affordable
when implemented with Z3 queries.
}
\section{Conclusion and future work}
\label{sec:conclusion}

We have presented a polynomial-time algorithm in Angluin's exact
learning framework using membership and equivalence for learning a finite 
union of rectilinear cubes over $\Z^d$ over any fixed dimension $d$.
By considering an additional subset oracle, the framework is also capable of
learning possibly infinite cubes in polynomial time over fixed dimensions, while
achieving a simpler and faster learning algorithm in practice.
At the core of our algorithms is the notion
of corners, whose search is efficiently implemented by a binary search. This
technique enables the introduction of auxiliary oracles, namely
the corner oracle (resp. maximal cube oracle) when a membership (resp. subset)
oracle is provided. While oracles for subset queries tend to be difficult to 
implement, this turns out not to be the case for our proposed application of
computing monadic decompositions of 
quantifier-free integer linear arithmetic formulas without modulo constraints.
We have demonstrated that our algorithms could be
successfully applied to this problem, substantially improving the existing 
algorithms.
\OMIT{
While our procedure are applied to and evaluated on the monadic decomposition
of quantifier-free integer linear arithmetic formulas, a direct
implementation of these auxiliary oracles in any decidable theory paves the
way to learning techniques on union of cubes in the real lattice
as a concept class.
}

We mention three future research directions. First, extensions to modulo
operations could be explored, by encoding periodicity on
$d$ additional coordinates and providing adequate oracles on the encoded
target.
A second direction consists in applying these learning techniques to the
   verification of systems by learning invariants which are monadically
   decomposable in a small number of cubes.
   Lastly, one promising direction to further improve our algorithms is to
investigate how to leverage if-then-else formula representations as used in
mondec$_1$ \cite{monadic-decomposition}, which could be exponentially more
succinct than formulas in DNF.

\OMIT{
Although some lower bounds were provided, it is not clear if the exponential
blow-up provided by our termination guarantees can be avoided. While the overshooting
issue could probably not be avoided, a formal lower bound is yet to be
explored either in the minimally adequate teacher setting, or alternatively
with the auxiliary corner oracle setting. On the other hand, introducing
the inclusion oracle is providing experimentally better performances of the
algorithm, at the expense of more expensive queries.

Inspired by the application to Presburger arithmetic, extensions to modulo
operators are yet to be explored.
A partial answer to these question is to shift the learning of
finite unions of ultimately periodic sequences of cubes in dimension~$d$ as
a concept $\concept\in\conceptCl$, to
the learning of another concept $\concept'\in\conceptCl$,
in dimension~$2d$ where $d$ extra coordinates are used to encode
the periodic behavior. We leave as future work the relationship and implementation
of oracles between the two concepts.
}


\bibliographystyle{splncs04}
\bibliography{refs}

\ifextended
   \clearpage 
\appendix

\begin{center}
    \LARGE \bfseries APPENDIX
\end{center}

\section{Proof of \cref{prop:membership}}
\begin{proof}
    We first check the following invariants satisfied by the main loop:
    $\vec{0}\leq \vec{v}\in X$ and for all $j \in \inter{1}{i}$, $\vec{v}+\vec{e}_j \notin X$.
    This property entails correctness at termination, namely
    $\vec{v}\in \MCorners{X}$ when exiting the main loop.
    At the beginning of the iteration, consider the set
    $C = \{(j,i')\in \inter{1}{n} \times \inter{1}{d}~|~ \overline{\vec{v}}_j[i'] > v[i'] \}$.
    Since $\vec{v}[i']$ can only increase, we have $C' \subseteq C$. We will
    prove now that the inclusion is strict by analyzing the inner body:
    \begin{itemize}
      \item After evaluating the first inner-loop, $k$ is defined such that
        $\vec{v}+(k/2)\cdot \vec{e}_i \in X$ and $\vec{v}+k\cdot \vec{e}_i\notin X$.Let us denote this
        value $k_0$ and notice that its computation required $\log k_0$ membership
        queries;
      \item The second inner-loop proceeds to a binary search to find
        some value $l\in\inter{k_0/2}{k_0}$ such that $\vec{v}+l\cdot \vec{e}_i\in X$ and
        $\vec{v}+(l+1)\vec{e}_i\notin X$.
        This computation required at most $\log k_0$ membership queries and
        ensured the existence of some
        $\overline{\vec{v}}_j$ such that $\overline{\vec{v}}_j[i] = l + \vec{v}[i]$.
    \end{itemize}
    Since $k_0 \leq 2\times \overline{\vec{v}}_j[i]$, the overall body of the main loop
    required at most $2\size{\overline{\vec{v}}_j[i]}$ where
    $(j,i)\in C\backslash C'$.
    The total number of membership queries is therefore bounded by
    $2\sum_{j=1}^n\sum_{i=1}^d \size{\overline{\vec{v}}_j[i]}$.
\end{proof}

\section{Proof of cube subtraction, lemma~\ref{lem:cubsub}}
\label{sec:proof-cubsub}
\begin{proof}
   \begin{itemize}
   \item We observe that
   $C'_2 = \Cube{\underline{v}'_2}{\overline{v}'_2} = C_1\cap C_2$
   such that for all $i\in[1;d]$,
   $\underline{v}'_2[j] = \max(\underline{v}_1[j], \underline{v}_2[j])$
   and
   $\overline{v}'_2[j] = \min(\overline{v}_1[j], \overline{v}_2[j])$.
   \item For the difference, we assume, without loss of generality, that
   $C_2 \subseteq C_1$ (by taking the intersection first).
   Remark that $v\notin C_2$ if, and only if,
   $\exists j: v[j] \geq \overline{v}_2[j]+1 \vee
    v[j] \leq \underline{v}_2[j] -1$, if, and only if,
   \[\exists! j:
    \forall k<j, \underline{v}_2[k] \leq v[k] \leq \overline{v}_2[k] \wedge
    \left(
    v[j] \geq \overline{v}_2[j]+1 \vee
    v[j] \leq \underline{v}_2[j] -1
    \right)
   \]
   Namely, the equations
   $
    \forall k<j, \underline{v}_2[k] \leq v[k] \leq \overline{v}_2[k] \wedge
    v[j] \geq \overline{v}_2[j]+1$
and 
   $
    \forall k<j, \underline{v}_2[k] \leq v[k] \leq \overline{v}_2[k] \wedge
    v[j] \leq \underline{v}_2[j] -1
    $, for $k\in[1;d]$, are mutually exclusive.
  \end{itemize}
\end{proof}

\section{Infinite concepts with M+EQ oracles}
\label{sec:infinitymemeq}
We adapt in this section the overshooting algorithm to the infinite concept
case of \cref{sec:infinity}, whose definitions are reused here.
In particular, we assume the equivalence oracle $\Psi_X$ to be defined on
the concept class $\conceptCl$ containing all finite unions of
unbounded cubes.
This overshooting variant is of theoretical interest as no implementation has
been considered in the submission.

For $r>0$, and a value $x\in\bbZ$, we write $\ext{x}{r}\in\bbZi$
defined by
\[
    \ext{x}{r} = \left\{\begin{aligned}
        \infty &\text{~if $x>=r$}\\
        -\infty &\text{~if $x<-r$}\\
        x&\text{~otherwise}
    \end{aligned}\right.
\]

The notation is extended coordinate-wise to any vector $v\in\bbZ^d$.
Furthermore, for $k\in\bbN$ we write
$\ball{r} = \{v\in\bbZ^d ~|~ \max_i |v[i]| \leq r\}$ for the ball of radius~$r$.

\begin{algorithm}
\begin{algorithmic}
  \Function{LearnCubesInfinity}{$\Phi_X,\Psi_X$} 
  \State Let $H\gets \emptyset$
  \State Let $V\gets \emptyset$
  \State Let $r \gets 1$
  \While{$H \neq X$}
    \State Let $v\in H \Delta X$
    \While{$v \notin \ball{r}$}
        \State $r\gets 2\cdot r$
    \EndWhile
    \State Let $v_1 = -\Call{findMaxCorner}{-v,-(X \Delta H)\cap \ball{r}}$
    \State Let $v_2 = \Call{findMaxCorner}{v,(X \Delta H\backslash
        \{ v~|~\exists v'_1\in V: v_1 \leq v'_1 \leq v  \})\cap \ball{r}}$
    \State $H \gets H \Delta \Cube{\ext{v_1}{r}}{\ext{v_2}{r}}$
    \State $V \gets V \uplus \{v_1\}$
  \EndWhile
  \EndFunction
\end{algorithmic}
\caption{Minimally adequate teacher for infinite concepts}
\label{alg:infinitymemq}
\end{algorithm}

Intuitively, the original learning procedure is applied to the target set $X$ restricted
to some ball of radius $2^t$, where $t$ may increase over time.
For a given $k\in[1;d]$, we define
$\underline{B}'_k = \{s\cdot 2^t~|~\exists s\in \{\pm 1\}
    \exists \alpha\in \underline{B}_k: |\alpha|>2^t\}\cup \underline{B}_k$
and
$\overline{B}'_k = \{s\cdot 2^t~|~\exists s\in \{\pm 1\}
    \exists \alpha\in \overline{B}_k: |\alpha|>2^t\}\cup \overline{B}_k$
Finally, $\intb'$ as the set of all possible union of cubes with
min and maximal corners in $(\underline{B}'_k)_k$
and $(\overline{B}'_k)_k$, respectively.

For every value in $\alpha \in \underline{B}_k$ there are at most $2\log |\alpha|$
additional values in $\underline{B}'_k$. A quadratic factor may consequently
appear when searching for corners.

In order for $H\in \intb'$ to hold as an invariant, similarly to~\cref{lem:inv},
we notice that any counterexample has to stay on the grid defined by $\intb'$.
This can be achieved for example by assuming the counterexample to be at minimum
distance (for the infinite norm) 

We summarize these assumptions in the following theorem:
\begin{theorem}
  \label{thm:infinity}
  Assume that for all $H\in \conceptCl\backslash \{X\}$,
  $\Psi_X(H)\in \Corners{X\Delta H}$.
  Then \Call{LearnCubesInfinity}{$\Phi_X,\Psi_X$} terminates in at most
  $O(\size{X})^{d}$ iterations where every iteration may require
  $O(\size{X}^2)$ membership queries and one equivalence query.
\end{theorem}

\fi

\end{document}